\documentclass[11pt,twoside]{article}
\usepackage{fancyhdr}
\usepackage[colorlinks,citecolor=blue,urlcolor=blue,linkcolor=blue,bookmarks=false]{hyperref}
\usepackage{amsfonts,epsfig,graphicx}
\usepackage{afterpage}
\usepackage{amsmath,amssymb,amsthm} 
\usepackage{fullpage}
\usepackage[T1]{fontenc} 
\usepackage{epsf} 
\usepackage{graphics} 
\usepackage{amsfonts,amsmath}
\usepackage[authoryear]{natbib} 
\usepackage{psfrag,xspace}
\usepackage{color,etoolbox}

\newtheorem{theorem}{Theorem}

\newcommand{\bulk}[1]{\mathcal{B}_{#1}(p_0)}
\newcommand{\truncnorm}[2]{V_{#1}(#2)}

\newcommand{\epstail}[1]{\mathcal{Q}_{#1}(p_0)}

\newcommand{\ssone}{n_1}
\newcommand{\sstwo}{n_2}

\newcommand{\func}{V}

\let\hat\widehat

\begin{document}

\begin{center} {\Large{\bf{Hypothesis Testing for High-Dimensional Multinomials: \\
\vspace{0.2cm}
A Selective Review}}}
\\

\vspace*{.3in}

{\large{
\begin{tabular}{ccccc}
Sivaraman Balakrishnan$^\dagger$  & ~~~~~~~~ &  Larry Wasserman$^{\dagger}$ \\
\end{tabular}

\vspace*{.1in}

\begin{tabular}{ccc}
Department of Statistics$^{\dagger}$ \\
\end{tabular}

\begin{tabular}{c}
Carnegie Mellon University, \\
Pittsburgh, PA 15213.
\end{tabular}

\vspace*{.2in}

\begin{tabular}{c}
{\texttt{\{siva,larry\}@stat.cmu.edu}}
\end{tabular}
}}

\vspace*{.2in}

\emph{In memory of Stephen E. Fienberg.}

\today

\vspace*{.2in}
\begin{abstract}
The statistical analysis of discrete data has been the subject of 
extensive statistical research dating back to the work of Pearson. 
In this survey we review some recently developed methods for
testing hypotheses about high-dimensional multinomials.
Traditional tests like the $\chi^2$-test and the likelihood ratio test
can have poor power in the high-dimensional setting. 
Much of the research in this area has focused on finding tests
with asymptotically Normal limits and developing (stringent) 
conditions under which tests have Normal limits. 
We argue that this perspective suffers from 
a significant deficiency: it can exclude many high-dimensional 
cases when --- despite having non-Normal 
null distributions --- carefully designed tests can have high power. 
Finally, we illustrate that taking a 
minimax perspective and considering refinements of this perspective can lead naturally to powerful and practical tests.
%But this rules out many important cases.
%The computer science literature contains new tests with better power
%but these are often not practical.
%We consider practical versions of these tests and we show that they have 
%higher power than traditional tests.
\end{abstract}
\end{center}

\section{Introduction}
Steve Fienberg was a pioneer in the development of
theory and methods for discrete data.
His textbook
(\cite{bishop1995holland})
remains one of the main references for the topic. Our focus in this review 
is on high-dimensional multinomial 
models where the number of categories $d$ can grow with, and possibly exceed the sample-size $n$.
Steve's paper (\cite{fienberg1973simultaneous}), written with with Paul Holland,
was one of the first to consider multinomial data
in the high-dimensional case.
In \cite{fienberg1976analysis},
Steve provided strong motivation for considering the high-dimensional setting: 

\vspace{0.2cm}

\emph{``The fact remains $\ldots$ that with the
extensive questionnaires of modern-day sample-surveys, and the
detailed and painstaking inventory of variables measured by biological
and social-scientists, the statistician is often faced with large
sparse arrays full of 0's and 1's, in need of careful analysis''.}

\vspace{0.2cm}

In this review we focus on hypothesis testing for high-dimensional
multinomials.  In the context of hypothesis testing, several works
(see for instance \cite{read88,holst72} and references therein) have
considered the high-dimensional setting. \cite{hoeffding1965}
(building on an unpublished result of Stein) showed that for testing
goodness-of-fit, in sharp contrast to the fixed-$d$ setting, in the
high-dimensional setting the likelihood ratio test can be dominated by
the $\chi^2$ test.  In traditional asymptotic testing theory, the
power of tests are often investigated at local alternatives which
approach the null as the sample-size grows. In the high-dimensional
setting, considering local alternatives, \cite{ivchenko78} showed that
neither the $\chi^2$ or the likelihood ratio test are uniformly
optimal. These results show some of the difficulties of using
classical theory to identify optimal tests in the high-dimensional
regime.

\cite{morris75} studied the limiting distribution of a
wide-range of multinomial test statistics and gave relatively
stringent conditions under which these statistics have asymptotically
Normal limiting distributions. In general, as we illustrate in our simulations, carefully designed
tests can have high power under much weaker conditions, even when the
null distribution of the test statistics are not Gaussian or $\chi^2$.
In many cases, understanding the limiting distribution of the test
statistic under the null is important to properly set the test
threshold, and indeed this often leads to practical tests. However, in
several problems of interest, including goodness-of-fit, two-sample
testing and independence testing we can determine practical
(non-conservative) thresholds by simulation.  In the high-dimensional
setting, rather than rely on asymptotic theory and local alternatives,
we advocate for using the minimax perspective and developing
refinements of this perspective to identify and study optimal tests.

Such a minimax perspective has been recently developed in a series of
works in different fields including statistics, information theory and
theoretical computer science and we provide an overview of some
important results from these different communities in this paper.

%In this paper we review some 
%recent advances in hypothesis testing for high-dimensional multinomial distributions.
%
% and 
%we refer the reader to the 
%
%The Cressie and Read, Holst (1972) already suggest unnatural to keep $d$
%fixed. 
%
%Hoeffding (1965) points out that LRT may no longer dominate
%$\chi^2$. Morris gave CLTs.
%\usepackage{natbib}

%Local power analysis, uniformly optimal tests do not exist (Ivchenko
%and Medvedev 1978). What statistic to use if $d$ grows and.

\section{Background}

Suppose that we have data of the form
$Z_1,\ldots, Z_n \sim P$
where $P$ is a $d$-dimensional multinomial and 
$Z_i\in\{1,\ldots, d\}$. We denote the probability mass function 
for $P$ by $p \in \mathbb{R}^d$. The set of all multinomials is then denoted by
\begin{equation}
\label{eqn:all_mults}
{\cal M} = \biggl\{ p = (p(1),\ldots, p(d)):\ p(j) \geq 0,\ {\rm for\ all\ }j,\ \sum_j p(j) =1 \biggr\}.
\end{equation}
We are interested in the case where $d$ can be large,
possibly much larger than $n$.
In this paper, we focus on two hypothesis testing problems, and defer a discussion of other
testing problems to Section~\ref{sec:discussion}.
The problems we consider are:
\begin{enumerate}
\item {\bf Goodness-of-fit testing: } In its most basic form, in
goodness-of-fit testing we are interested in testing the fit of the
data to a fixed distribution $P_0$.  Concretely, we are interested
in distinguishing the hypotheses:
$$
H_0: P=P_0\ \ \ {\rm versus}\ \ \ H_1: P\neq P_0.
$$
\item {\bf Two-sample testing: } In two-sample testing we observe
$$
Z_1,\ldots, Z_{\ssone} \sim P,\ \ \ 
W_1,\ldots, W_{\sstwo} \sim Q.
$$
In this case, the hypotheses of interest are
$$
H_0: P=Q\ \ \ {\rm versus}\ \ \ H_1: P\neq Q.
$$
\end{enumerate}
{\bf Notation: } Throughout this paper, we write $a_n\asymp b_n$
if both $a_n/b_n$ is bounded away from 0 and $\infty$ for all large $n$.

\subsection{Why Hypothesis Testing?}
As with any statistical problem, there are many inferential tasks
related to multinomial models:
estimation, constructing confidence sets, Bayesian inference, prediction and hypothesis testing,
among others.

Our focus on testing in this paper is not meant to
downplay the importance of these other tasks.
Indeed, many would argue that hypothesis testing
has received too much attention: over-reliance
on hypothesis testing is sometimes cited as one of the
causes of the reproducibility crisis.
However, there is a good reason for studying hypothesis testing.
When trying to understand the theoretical behavior
of statistical problems in difficult cases --- such as in high dimensional models ---
hypothesis testing provides a very clean, precise framework.

%For example, as we shall see later in this paper,
%some high dimensional multinomial models
%are intrinsically more difficult to test than others.
%This point is easy to make precise in the hypothesis testing framework, and we do so
%by quantifying what we refer to as the \emph{local minimax rate}.
Hypothesis testing is a good starting point for theoretical
investigations into difficult statistical models.
As an example,
in Section \ref{section::local}
we will see that the power of goodness-of-fit tests
can vary drastically depending on where the null sits in the simplex, a
phenomenon that we refer to as \emph{local minimaxity}. 
This local minimax phenomenon is very clear and easy to precisely capture
in the hypothesis testing framework.

\subsection{Minimax and Local Minimax Testing}
In traditional asymptotic testing theory, local measures of performance are often used
to assess the performance of various hypothesis tests.
%The are many ways to do so but the two most common approaches
%are (i) local measures of performance and
%(ii) global measures, specifically, minimax power.
In the local approach,
one typically examines the power at a sequence of alternatives at 
$\theta_0 + C/\sqrt{n}$ where $\theta_0$ denotes the null value of the parameter.
The local approach is well-suited to well-behaved, low dimensional models.
It permits very precise power comparisons for distributions which are
close to the null hypothesis.
Generally, the tools for local analysis are tied to
ideas like contiguity and asymptotic Normality and 
in the high dimensional setting these tools
often break down. Furthermore, as we discussed earlier, results of 
\cite{ivchenko78} suggest that the local perspective does not provide
a clear overall picture in the high-dimensional case.
%The traditional limiting distributional results don't hold.
%And the behavior of the power can be very sensitive to the direction
%of the local alternative which then means that the local
%perspective does not give a good, overall picture of how the problem behaves.

For these reasons, we will use the minimax perspective. For goodness-of-fit testing, 
we can refine the minimax perspective to obtain results that also have a local nature, but in a very different sense than the local results described above.
Formally, a test is a map from the samples to $\{0,1\}$. We 
%(We shall see that the minimax perspective can also have a local nature,
%but in a different sense than the approach described above.)
let $\Phi_n$ denote all level $\alpha$ tests, i.e. $\phi\in\Phi_n$ if
$\phi(Z_1,\ldots, Z_n)\in\{0,1\}$ and
$$
\sup_{P\in {\cal P}_0}P^n(\phi=1) \leq \alpha,
$$
where $\mathcal{P}_0$ denotes the (possibly composite) collection of possible null distributions.
Let
$$
{\cal M}_\epsilon = \Bigl\{p:\ d({\cal P}_0,p) \geq \epsilon\Bigr\}
$$
where
${\cal P}_0$ is the set of null distributions,
$d({\cal P}_0,p) = \inf_{q\in {\cal P}_0}d(q,p)$
and $d(p,q)$ is some distance.
The maximum type II error of a test $\phi\in\Phi_n$ is
\begin{equation*}
R_{n,\epsilon}(\phi,\mathcal{P}_0)=\sup_{P \in {\cal M}_\epsilon} P^n(\phi=0).
\end{equation*}
The {\em minimax risk} is
\begin{equation*}
R_{n,\epsilon}^\dagger(\mathcal{P}_0) = \inf_{\phi\in\Phi_n} R_{n,\epsilon}(\phi,\mathcal{P}_0).
\end{equation*}
A test $\phi\in\Phi_n$ is minimax optimal if
$R_{n,\epsilon}(\phi)=R_{n,\epsilon}^\dagger(\mathcal{P}_0).$
%We often 
%(In practice, we might only have that
%$R_{n,\epsilon}(\phi)\approx R_{n,\epsilon}^\dagger$.)
It is common to study the minimax risk via a 
coarse lens by studying instead the 
\emph{minimax separation}, also called the 
{\em critical radius}. 
The minimax separation $\epsilon_n$
is defined by
\begin{equation*}
\epsilon_n(\mathcal{P}_0) = \inf\Bigl\{ \epsilon:\ R_{n,\epsilon}^\dagger(\mathcal{P}_0) \leq 1/2 \Bigr\}
\end{equation*}
which is the smallest $\epsilon$ such that the power is non-trivial. 
The choice of 1/2 is not important and any sufficiently small, non-zero number will suffice. 

We need to choose a distance $d$.
We will focus on the total variation (TV) distance
defined by
\begin{equation*}
{\rm TV}(P,Q) = \max_A | P(A)-Q(A)|
\end{equation*}
where the maximum is over all events $A$.
The reason we use total variation distance
is because it has a clear probabilistic meaning and is invariant to natural transformations \citep{devroye85}:
if ${\rm TV}(P,Q) = \epsilon$ then
$|P(A) - Q(A)| \leq \epsilon$ for every event $A$.
The total variation distance is equivalent to the $L_1$ distance:
\begin{equation*}
{\rm TV}(P,Q) = \frac{1}{2}\sum_j | p(j)-q(j)| \equiv \frac{1}{2}\|p-q\|_1
\end{equation*}
where
$p$ and $q$
are the probability functions corresponding to the distributions $P$ and $Q$.
Other distances, such as $L_2$, Hellinger and Kullback-Leibler can be used too, but in some cases can be less interpretable and in other cases lead to trivial minimax rates. We revisit the choice of metric in Section~\ref{sec:other_metrics}.

Typically we characterize the minimax separation by providing upper and lower bounds on it: upper bounds are obtained by analyzing the minimax separation for practical tests, while lower bounds are often obtained via an analysis of the likelihood ratio test for carefully constructed pairs of hypotheses (see for instance the pioneering work of Ingster and co-authors \cite{ingster03,ingster1997adaptive}).

\vspace{0.2cm}

\noindent {\bf The Local Minimax Separation: } Focusing on the problem of goodness-of-fit testing with a simple null, we observe that the minimax separation $\epsilon_n(p_0)$ is a function of the null distribution. Classical work in hypothesis testing \cite{ingster03,ingster1997adaptive} has focused on characterizing the global minimax separation, i.e. in understanding the quantity:
\begin{equation}
\label{eqn:gm}
\epsilon_n = \inf\Bigl\{ \epsilon:\ \sup_{p_0 \in \mathcal{M}} R_{n,\epsilon}^\dagger(p_0) \leq 1/2 \Bigr\},
\end{equation}
where $\mathcal{M}$ is defined in~\eqref{eqn:all_mults}. In typical non-parametric problems, the local minimax risk and the global minimax risk match up to constants and this has led researchers in past work to focus on the global minimax risk.

Recent work by \cite{valiant2017automatic} showed that for goodness-of-fit testing in the TV metric for high-dimensional multinomials, the critical radius can vary considerably as a function of the null distribution $p_0$. In this case, the local minimax separation, i.e. $\epsilon_n(p_0)$ provides a much more refined notion of the difficulty of the goodness-of-fit problem. \cite{valiant2017automatic} further provided a locally minimax test, i.e. a test that is nearly-optimal for \emph{every possible null distribution}. 

Developing refinements to the minimax framework in problems beyond goodness-of-fit testing is an active area of research. 
For the problem of two-sample testing for multinomials, a recent proposal appears in the work of \cite{acharya12}.  Other works
developing a local perspective in testing and estimation include \cite{donoho94,gold11,caisinica,cha15,wei17}.

%We will place no restrictions on the distributions.
%This is in contrast to much of the literature on high dimensional multinomials
%where extra assumptions are often imposed so that
%limiting distributions can be obtained.
%Instead, we take an assumption-free approach which
%means that the familar tools of asymptotic theory are not helpful for
%the analysis.
%In each problem,
%we are interested in the minimax separation $\epsilon_n$.
%We also identify tests that achieves this power.
%We then provide some simulation studies that compare the power of several tests.

\section{Goodness-of-Fit}
\label{section::goodness}

Let $Z_1,\ldots, Z_n \sim P$
where $Z_i \in \{1,\ldots, d\}$.
Define the vector of counts
$(X_1,\ldots,X_d)$ 
where $X_j = \sum_{i=1}^n \mathbb{I}(Z_i =j)$.
We consider testing the simple null hypothesis:
$$
H_0: P=P_0\ \ \ {\rm versus}\ \ \ H_1: P \neq P_0.
$$
The most commonly used test statistics are the chi-squared statistic,
\begin{align}
\label{eqn:oldchisq}
T_{\chi^2} = \sum_{j=1}^d \frac{(X_j - n p_0(j))^2}{n p_0(j)},
\end{align}
and the likelihood ratio test (LRT)
\begin{align}
\label{eqn:lrt}
T_{\text{LRT}} = \sum_{j=1}^d \hat p(j) \log\left(\frac{\hat p(j)}{p_0(j)}\right),
\end{align}
where $\hat p(j) = X_j/n$. See the book of \cite{read88} for a variety of other popular multinomial goodness-of-fit tests. 
As we show in Section~\ref{sec:sims},
when $d$ is large,
these tests can have poor power. 
In particular, they are not minimax optimal.
Much of the research on tests in the large $d$ setting
has focused on establishing conditions under which
these statistics have convenient limiting distributions (see for instance \cite{morris75}).
Unfortunately, these conditions are generally not testable,
and they rule out many interesting cases, when test statistics despite not having a 
convenient limiting distribution can have high power (see Section~\ref{sec:sims}).

\subsection{Globally Minimax Optimal Tests}
The global minimax separation rate was characterized in 
the works of \cite{paninski2008coincidence,valiant2017automatic}. In particular, these works show that the global minimax separation rate (see~\eqref{eqn:gm}) is given by:
\begin{align}
\label{eqn:gm_mult}
\epsilon_n \asymp \frac{d^{1/4}}{\sqrt{n}}.
\end{align}
This implies, surprisingly, that we can have non-negligible power even when $n \ll d$. In the regime when $n \asymp \sqrt{d}$ most categories of the multinomial are unobserved, but we can still distinguish \emph{any} multinomial from alternatives separated in $\ell_1$ with high power. 
In start contrast, it can be shown that the minimax estimation rate in the $\ell_1$ distance, is $\sqrt{d/n}$ which is much slower than the testing rate. This is a common phenomenon: hypothesis testing is often easier than estimation.
%For example, it suffices that
%$n \approx \sqrt{d}$.

An important fact, elucidated by the minimax perspective, is that none of the traditional tests are minimax. 
A very simple minimax test, from 
\cite{balakrishnan2017hypothesis}
is the truncated $\chi^2$ test defined by
\begin{equation}\label{eq::minimax-test}
T_{\text{trunc}} = \sum_j \frac{(X_j - np_0(j))^2 - X_j}{ \max\{p_0(j), \frac{1}{d}\}}.
\end{equation}
The $\alpha$ level critical value $t_\alpha$
is defined by
$$
t_\alpha(p_0) = \inf\Bigl\{t:\ P_0^n(T> t) \leq \alpha \Bigr\}.
$$
Normal approximations or $\chi^2$ approximations cannot be used to find $t_\alpha$
since the asymptotics are not uniformly valid over ${\cal M}$.
However, the critical value $t_\alpha$ for the test
can easily be found by simulating from $P_0$.
In Section 
\ref{sec:sims}
we report some simulation studies that illustrate the gain in power
by using this test.

\subsection{Locally Minimax Optimal Tests}
\label{section::local}
%A subtle issue --- first discovered in the theoretical computer science literature ---
%is that the minimax rate can depend on the null $p_0$.
%We call this the {\em local minimax rate}.
%The local minimax rate was discovered by
%\cite{valiant2017automatic}
%(who refer to it as {\em instance optimality}).
%Moreover, they designed a new test that achieve this optimal rate.
In goodness-of-fit testing, some nulls are easier to test than others.
For example, when $p_0$ is uniform the local minimax risk is quite large and scales as in~\eqref{eqn:gm_mult}. However, when $p_0$ is sparse the problem effectively behaves as a much lower dimensional multinomial testing problem and the minimax separation can be much smaller.
This observation has important practical consequences.
Substantial gains in power can be achieved,
by adapting the test to the shape of the null distribution $p_0$.

Roughly, \cite{valiant2017automatic}
showed that
the local minimax rate is given by
$$
\epsilon_n(p_0) \asymp \sqrt{\frac{\func(p_0)}{n}}
$$
for a functional $\func$ that depends on $p_0$ as,
$$
\func(p_0) \approx \|p_0\|_{2/3} = \left(\sum_{j=1}^d p_0^{2/3}(j)\right)^{3/2}.
$$
We provide a more precise statement in the Appendix.
The fact that the local minimax rate depends on the $2/3$ norm
is certainly not intuitive and is an example
of the surprising nature of the results in the world of high dimensional multinomials.
When $p_0$ is uniform the 2/3-rd norm is maximal and takes the value
$\|p_0\|_{2/3} = \sqrt{d}$
whereas when $p_0 = (1,0,\ldots, 0)$ the 2/3-rd norm is much smaller, i.e. 
$\|p_0\|_{2/3} = 1$.
This means that a test tailored to $p_0$
can have dramatic gains in power.

\cite{valiant2017automatic}
constructed 
a test that achieves
the local minimax bound.
To describe the test we need a few definitions.
First, without loss of generality, assume that
$p_0(1) \geq p_0(2)\geq \cdots \geq p_0(d)$.
Let $\sigma\in [0,1]$ and define the tail and bulk by
$$
{\cal Q}_\sigma(p_0) = \Bigl\{ i:\ \sum_{j=i}^d p_0(j) \leq \sigma \Bigr\}
$$
and
$$
{\cal B}_\sigma(p_0) = \Bigl\{ i > 1:\ i\notin {\cal Q}_\sigma(p_0)\Bigr\}.
$$
The test is
$\phi = \phi_1 \vee \phi_2$
where
$\phi_1 = I( T_1(\sigma) > t_1)$,
$\phi_2 = I( T_2(\sigma) > t_2)$,
\begin{align*}
T_1(\sigma) &= 
\sum_{j \in {{\cal Q}_\sigma}} (X_j - np_0(j)),
~~~~~~~~~~~~~~~t_{1}(\alpha,\sigma) =  \sqrt{ \frac{n P_0({{\cal Q}_\sigma}) }{\alpha}}, \\
T_2(\sigma) &= 
\sum_{j\in {{\cal B}_\sigma}} \frac{(X_j - n p_0(j))^2 - X_j}{p^{2/3}_0(j)},
~~~~~t_{2}(\alpha,\sigma) = \sqrt{ \frac{\sum_{j \in {\cal B}_\sigma} 2n^2 p_0(j)^{2/3}}{\alpha}}. 
\end{align*}
The test may appear to be somewhat complicated but
all the quantities are easy to compute. Furthermore, in practice the thresholds are easily computed by simulation. Other near-local minimax tests for testing multinomials appear in \cite{diakonikolas2016new,balakrishnan2017hypothesis}.

A problem with the above test is that there
is a tuning parameter $\sigma$.
\cite{valiant2017automatic}
suggested using
$\sigma = \epsilon/8$.
While this choice is useful for theoretical analysis it 
is not useful in practice as
it would require knowing how far $P$ is from $P_0$
if $H_0$ is false. 
In \cite{balakrishnan2017hypothesis} we propose ways
to select the tuning parameter $\sigma$ in a data-driven fashion. 
For instance, one might consider a Bonferroni corrected test.
More precisely, let $\Sigma = \{\sigma_1,\ldots, \sigma_N\}$
be a grid of values for $\sigma$.
Let $\phi_j$ be the test using tuning parameter $\sigma_j$
and significance level $\alpha/N$.
We then use the Bonferroni corrected test
$\phi = \max_j\{ \phi_j\}$. It can be shown that,
if $\Sigma$ is chosen carefully,
there is only a small loss of power
from the Bonferroni correction.

\subsection{Implications for Continuous Data}

Although the focus of this paper is on discrete data,
we would like to briefly mention the fact that these results
have implications for continuous data.
This discussion is based on
\cite{balakrishnan2017hypothesis}.

Suppose that $X_1,\ldots,X_n \sim p$
where $p$ is a density on $[0,1]$.
We want to test
$H_0: p=p_0$.
As shown in \cite{lecam73} and \cite{barron89},
the power of any test over the set
$\{p:\ {\rm TV}(p,p_0)> \epsilon\}$
is trivial unless we add further assumptions.
For example, suppose we restrict attention to densities $p$ that satisfy
the Lipschitz constraint:
\begin{equation*}
  |p(y) - p(x)| \leq L |x-y|.
\end{equation*}
In this case,
\cite{ingster1997adaptive}
showed that, when $p_0$ is the uniform density,
the minimax separation rate is
$\epsilon_n \asymp n^{-2/5}$.
The optimal rate can be achieved
by binning the data and using a $\chi^2$ test.
However,
if $p_0$ is not uniform and
the Lipschitz constant $L$ is allowed to grow with $n$,
\cite{balakrishnan2017hypothesis}
showed that the local minimax rate is
\begin{align*}
\epsilon_n(p_0) \approx \left( \frac{\sqrt{L_n} T(p_0)}{n}\right)^{2/5}
\end{align*}
where $T(p_0)\approx \int \sqrt{p_0(x)}dx$.
This rate can be achieved
by using a very careful, adaptive binning procedure,
and then invoking the test
from Section \ref{section::local}.
The proofs make use of some of the tools for high dimensional multinomials
described in the previous sections.

The main point here is that the theory for multinomials has
implications for continuous data.
It is worth noting that
\cite{fienberg1973simultaneous}
was one of the first papers to explicitly link
the high dimensional multinomial problem to
continuous problems.

\subsection{Testing in Other Metrics}
\label{sec:other_metrics}

A natural question is to characterize the dependence of the local
minimax separation and the local minimax test on the choice of
metric. We have focused thus far on the TV metric.
The paper
\cite{daska18} considers goodness-of-fit testing in other
metrics. They provide results on the global minimax separation for the
Hellinger, Kullback-Leibler and $\chi^2$ metric.  In particular, they
show that while the global minimax separation is identical for
Hellinger and TV, this separation is infinite for the Kullback-Leibler
and $\chi^2$ distance because these distances can be extremely
sensitive to small perturbations of small entries of the multinomial.

In forthcoming work \citep{forth} we characterize the \emph{local} minimax rate in the Hellinger metric for high-dimensional multinomials as well as for continuous distributions. The optimal choice of test, as well as the local minimax separation can be sensitive to the choice of metric and (in the continuous case) to the precise nature of the smoothness assumptions. 

Despite this progress, developing a comprehensive theory for minimax testing of high-dimensional multinomials in general metrics, which provides useful practical insights, remains an important open problem.

\subsection{Composite Nulls and Imprecise Nulls}

Now we briefly consider the problem of goodness-of-fit testing
for a composite null.
Let ${\cal P}_0\subset {\cal M}$ be a subset of multinomials and consider testing
\begin{equation*}
H_0: P\in {\cal P}_0\ \ \ {\rm verus}\ \ \ H_1: P\notin {\cal P}_0.
\end{equation*}
A complete minimax theory for this case is not yet available, but many
special cases have been studied.  In particular, the work of
\cite{acharya15}, provides some results for testing monotonicity,
unimodality and log-concavity of a high-dimensional multinomial. Here,
we briefly outline a general approach due to \cite{acharya15}.

A natural approach to hypothesis testing with a composite null is to
split the sample, estimate the null distribution using one sample, and
then to test goodness-of-fit to the estimated null distribution using
the second sample. In the high-dimensional setting, we cannot assume
that our estimate of the null distribution is very accurate
(particularly in the TV metric). However, as highlighted in
\cite{acharya15} even in the high-dimensional setting we can often
obtain sufficiently accurate estimates of the null distribution in the
$\chi^2$ distance. This observation motivates the study of the
following two-stage approach.
Use half the data to get an estimate $\hat p_0$ assuming $H_0$ is true.
Now use the other half to test the imprecise null of the form:
\begin{equation*}
H_0: d_1(p,\hat p_0) \leq \theta_n \ \ \ {\rm verus}\ \ \ H_1: d(p,\hat p_0) \geq \epsilon_n.
\end{equation*}
Here, $\hat p_0$ is treated as fixed.
\cite{acharya15} refer to this as ``robust testing''
but they are not using the word robust in the usual sense.
This imprecise null testing problem has been studied for a variety of metric 
choices in \cite{valiant11,acharya15} and \cite{daska18}. 

Towards developing practical tests for general composite nulls, an
important open problem is to provide non-conservative methods for
determining the rejection threshold for imprecise null hypothesis tests. In
the high-dimensional setting we can no longer rely on limiting
distribution theory, and it seems challenging to develop simulation
based methods in general. Some alternative proposals have been
suggested for instance in \cite{berger1994p}, but warrant further
study in the high-dimensional setting.

More specific procedures can be constructed based on the structure of
${\cal P}_0$, and important special cases of composite null testing 
such as two-sample testing and 
independence testing have been studied in the literature. We turn our attention towards two-sample testing next. 

%As an example, we will consider the two sample problem
%in a later section.

\section{Two Sample Testing}
In this case the data are
$$
Z_1,\ldots,Z_{\ssone} \sim P,\ \ \ \ \ 
W_1,\ldots,W_{\sstwo} \sim Q
$$
and the hypotheses are
$$
H_0: P=Q\ \ \ {\rm versus}\ \ \ H_1: P\neq Q.
$$
Let $X$ and $Y$ be the corresponding vectors of counts.
First, suppose that $\ssone=\sstwo := n$.
In this case,
\cite{chan2014optimal}
showed that 
the minimax rate is
\begin{equation}
\epsilon_n \asymp \max \Biggl\{
\frac{d^{1/2}}{n^{3/4}},\
\frac{d^{1/4}}{n^{1/2}} \Biggr\}.
\end{equation}
The second term in the maximum is identical to the goodness-of-fit rate. 
In the high-dimensional case $d\geq n$, the minimax separation rate is
$\frac{d^{1/2}}{n^{3/4}}$, which is strictly slower than the goodness-of-fit rate. This highlights that, in contrast to the low-dimensional setting, there can be a price to pay for testing when the null distribution is not known precisely (i.e. for testing with a composite null).

\cite{chan2014optimal} also showed that the (centered) $\chi^2$ statistic
\begin{equation}
\label{eqn:chisq_stat}
T = \sum_j \frac{(X_j - Y_j)^2 - X_j - Y_j}{X_j + Y_j}
\end{equation}
is minimax optimal. The critical value $t_\alpha$ can be obtained using the usual
permutation procedure \citep{lehmann06}.
It should be noted, however, that the theoretical results
do not apply to the data-based permutation cutoff but, rather, to a
cutoff based on a loose upper bound on the variance of $T$. In the remainder of this section we discuss some extensions:
\begin{enumerate}
\item {\bf Unequal Sample Sizes: } The problem of two-sample testing with unequal sample sizes has been considered in \cite{diakonikolas2016new} and \cite{bhattacharya2015testing}, and we summarize some of their results here.
%Now suppose that the sample sizes are unequal.
Without loss of generality, assume that $\ssone \geq \sstwo$.
%\cite{
The minimax rate is
\begin{equation}
\epsilon_n \asymp \max
\Biggl\{
\frac{d^{1/2}}{\ssone^{1/4}\sstwo^{1/2}},\
  \frac{d^{1/4}}{\sstwo^{1/2}} \Biggr\}.
\end{equation}
Note that this rate is identical to the minimax goodness of fit rate
from Section \ref{section::goodness} when $\ssone \geq d$.
This makes sense since, when $\ssone$ is very large, $P$ can be estimated to high precision
and we are essentially back in the simple goodness-of-fit setting.

From a practical perspective, while the papers \citep{diakonikolas2016new,bhattacharya2015testing} propose tests that are near-minimax they are not tests in the usual statistical sense.
They contain a large number of unspecified constants
and it is unclear how to choose the test threshold in such a way that
the level of the test is $\alpha$. We could choose the constants somewhat loosely then use
the permutation distribution to get a level $\alpha$ test, but it
is not known if the resulting test is still minimax, highlighting an important gap between theory and practice.

%Developing a practical version of the test and showing that it is minimax is
%the subject of current work.
%In our simulations, following \cite{bhattacharya2015testing} we use the classical two-sample $\chi^2$ statistic
%\begin{equation}
%  T = \sum_j \frac{(m X_j - n Y_j)^2 - (m^2 X_j+ n^2 Y_j)}{n^{3/2}m (X_j+Y_j)}
%  \end{equation}
%rather than the actual minimax test, and we report some simulations with this statistic in Section~\ref{sec:sims}.
%The usual $\chi^2$ test statistic for this problem
%is
%\begin{equation}
%  T = \sum_j
%  \left[ \frac{(X_j - \pi C(j))^2}{\pi (1-\pi)C(j)} -1\right]
%\end{equation}
%where
%$\pi = n/(n+m)$ and
%$C(j) = X_j+Y_j$.
%(The statistic has been centered to have mean 0 under $H_0$.)
%We will compare the performance of the $\chi^2$ test
%to the minimax bound by simulation in Section
%\ref{sec:sims}.

%{\bf Note: I think these two tests are the same}

\item {\bf Refinements of Minimaxity: } In two-sample testing, unlike in goodness-of-fit testing, it is less clear how precisely to define the local minimax separation. Roughly, we would expect that distinguishing whether two samples are drawn from the same distribution or not 
should be more difficult if the distributions of the samples are nearly uniform, than if the distributions are concentrated on a small number of categories. Translating this intuition into a satisfactory refined minimax notion is more involved than in the goodness-of-fit problem.

One such refinement appears in the work of \cite{acharya12}, who restrict attention to so-called ``symmetric'' test statistics (roughly, these statistics are invariant to re-labelings of the categories of the multinomial). Their proposal is in the spirit of classical notions of adaptivity and oracle inequalities in statistics (see for instance \cite{sara16,donoho96} and references therein), where they benchmark the performance of a test/estimator against an oracle test/estimator which is provided some side-information about the local structure of the parameter space. Concretely, they compare the separation achieved by $\chi^2$-type tests to the minimax separation achieved by an oracle that knows the distributions $P, Q$ up to a permutation.

%They then benchmark the performance of a 

%over the parameter space (see, for instance, the discussion and references in
%Donoho et al. [DJKP95]). In such situations, it can be fruitful to benchmark the risk of an estimator
%against that of a so-called oracle estimator that is provided with side-information about the local
%structure of the parameter space. Such a benchmark can be used to show that a given estimator is
%adaptive, in the sense that even though it is not given side-information about the problem instance,
%it is able to achieve lower risk for ?easier? problems (e.g., see the papers [Can06, Kol11, CL11]
%for results of this type).1

\item {\bf Testing Continuous Distributions: } Finally, we note that results for two-sample testing of high-dimensional multinomials have implications for two-sample testing in the continuous case. The recent paper of \cite{ariascastro16} building on work by \cite{ingster1997adaptive}, considers two-sample testing of smooth distributions by reducing the testing problem to an appropriate high-dimensional multinomial testing problem.
Somewhat surprisingly, they observe that at least under sufficiently strong smoothness assumptions, the minimax separation rate for two-sample testing matches that of goodness-of-fit testing. 

\end{enumerate}
\section{Simulations}
\label{sec:sims}
\begin{center}
\begin{figure}[h]
\begin{tabular}{cc}
\includegraphics[scale=0.4]{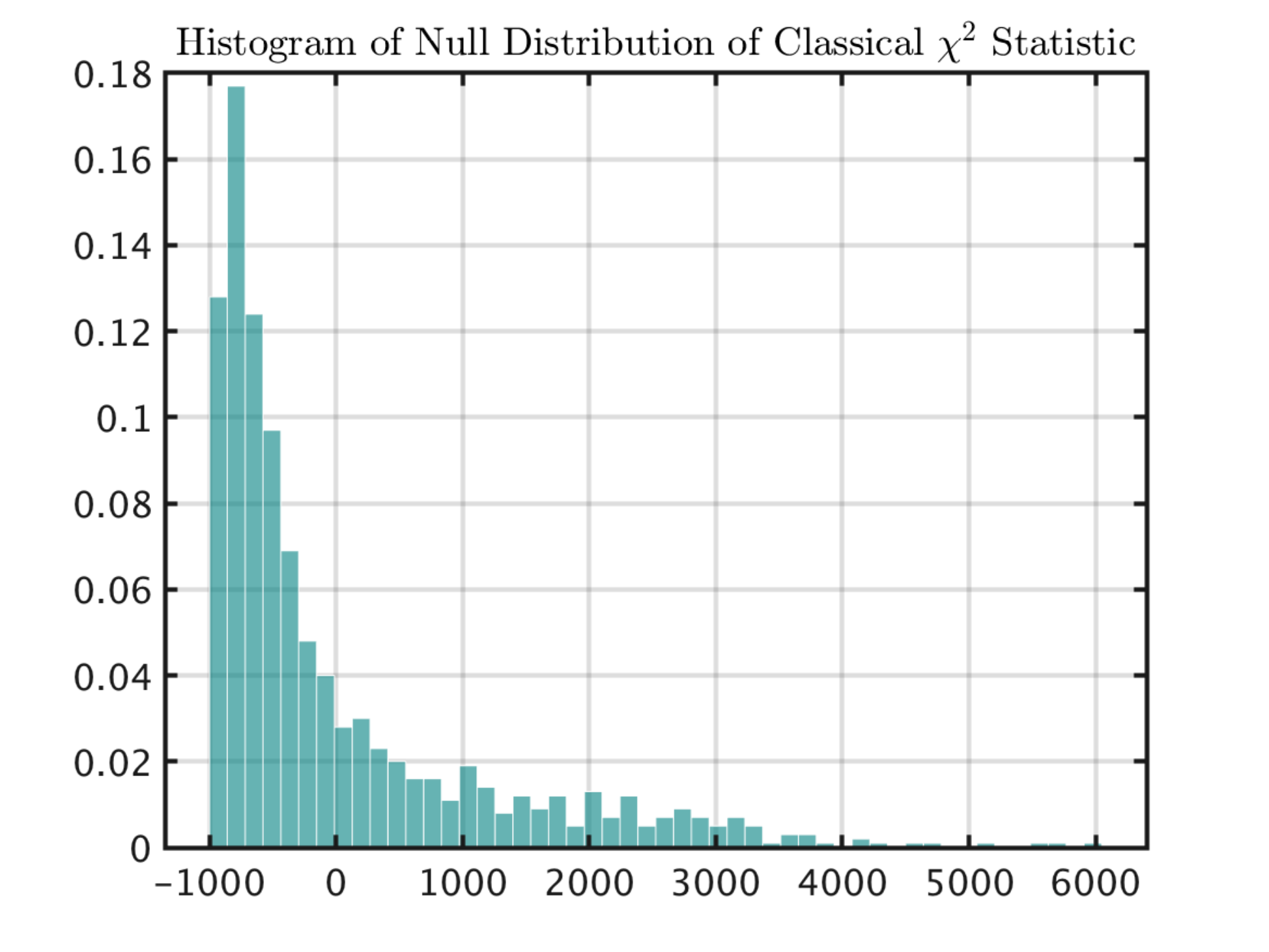} &~~~ \includegraphics[scale=0.4]{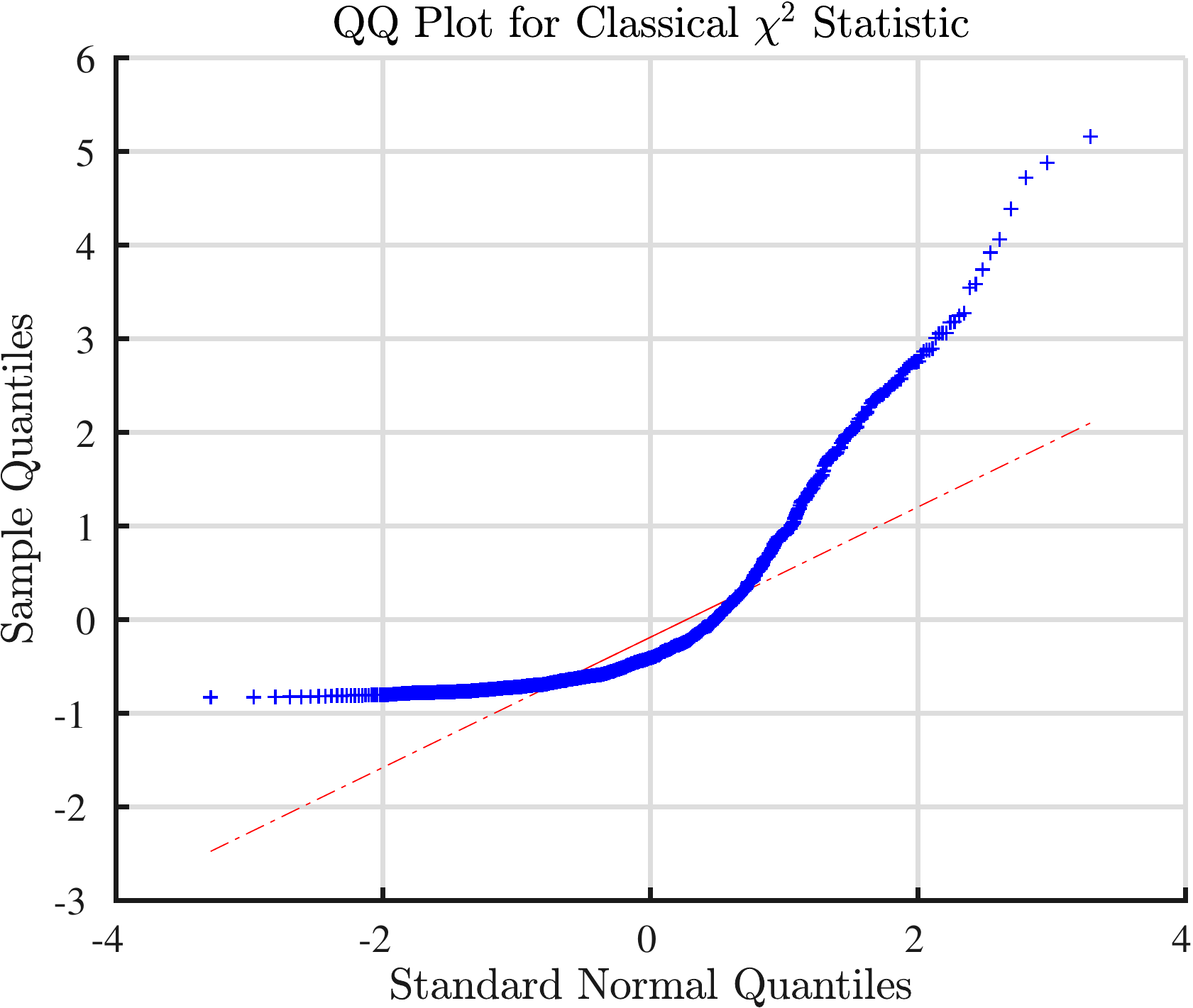} \\
\includegraphics[scale=0.4]{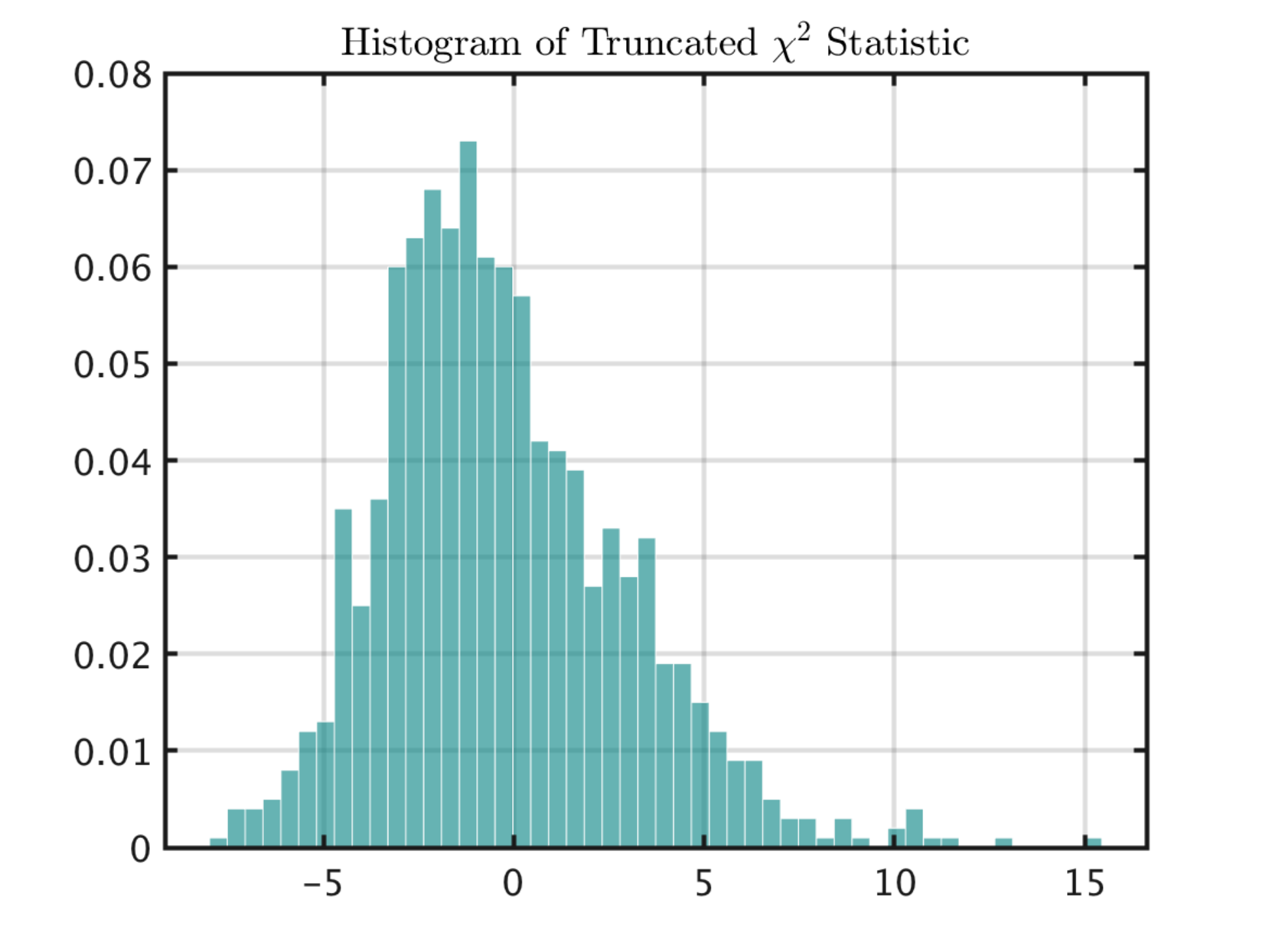} &~~~ \includegraphics[scale=0.4]{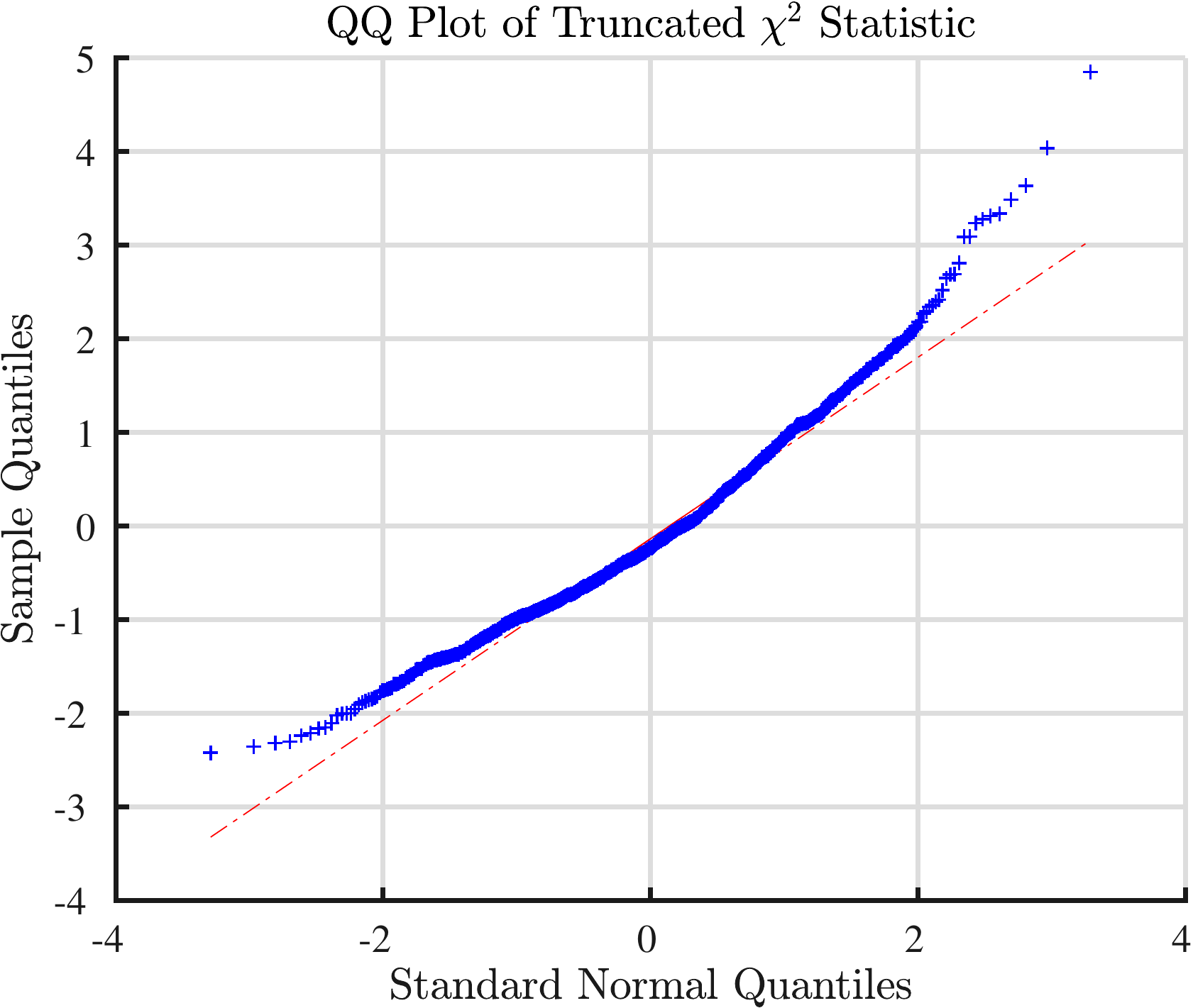}
\end{tabular}
\caption{A plot of the distribution of the 
classical and truncated $\chi^2$ test statistics under a power law null distribution, in the high-dimensional setting where $n = 400, d = 1000$,
obtained via simulation.}
\label{fig:limit}
\end{figure}
\end{center}
In this section, we report some simulation studies performed to illustrate that tests can have high power even when their limiting distributions are not Gaussian and to illustrate the gains from using careful modifications to classical tests for testing high-dimensional multinomials.

\subsection{Limiting distribution of test statistics} 
One of the main messages of our paper is that tests can have high-power even in regimes where their null distributions are not Gaussian (or more generally well-behaved). As a result, restricting attention to regimes where the null distribution of a test statistic is well-behaved can be severely limiting.

As an illustration, we consider goodness-of-fit testing where the null distribution is a power law, i.e. 
we take $p_0(i) \propto 1/i$. We will consider a high-dimensional setting where $d = 1000$ and $n = 400$. In this setting, as we will show in Section~\ref{sec:simgof}, the $\chi^2$-statistic performs poorly but the truncated statistic in~\eqref{eq::minimax-test} has high power. However, as illustrated in Figure~\ref{fig:limit} in this high-dimensional regime the limiting distributions of both statistics are quite far from Normal. We also observe that the classical $\chi^2$ statistic has a huge variance, which explains its poor power and motivates our introduction of the truncated $\chi^2$ statistic. The truncated $\chi^2$ statistic has a high-power, and achieves the minimax rate for goodness-of-fit testing, and as illustrated in this simulation has a much better behaved distribution under the null.

\subsection{Testing goodness-of-fit}
\label{sec:simgof}
In this section, we compare the performance of several goodness-of-fit test statistics. Throughout we take $n = 400$ and $d = 1000$. Closely related simulations appear in our prior work \cite{balakrishnan2017hypothesis}.
In particular, we compare the classical $\chi^2$ statistic in~\eqref{eqn:oldchisq}, the likelihood-ratio test in~\eqref{eqn:lrt}, the truncated $\chi^2$ statistic in~\eqref{eq::minimax-test} and the two-stage locally minimax 2/3rd and tail test described in Section~\ref{section::local}, with the $\ell_1$ and $\ell_2$ tests given as
\begin{align*}
T_{\ell_1} = \sum_{i=1}^d |X_i - np_0(i)|,~~~\text{and}~~~T_{\ell_2} = \sum_{i=1}^d  (X_i - np_0(i))^2.
\end{align*}
We examine the power of these tests under various alternatives:
\begin{enumerate}
\item {\bf Minimax Alternative: } We perturb each entry by an amount proportional to $p_0(i)^{2/3}$ with a randomly chosen sign. This is close to the worst-case perturbation used in \cite{valiant2017automatic} in their proof of the local-minimax lower bound.  
\item {\bf Uniform Dense Alternative: } We perturb each entry of the null distribution by a scaled 
Rademacher random variable. 
(A Rademacher random variable takes values $+1$ and $-1$ with equal probability.) 
\item {\bf Sparse Alternative: } In this case, we essentially only perturb the first two entries of the null multinomial.  
We increase the two largest entries of the multinomial and then re-normalize the resulting distribution, this results in a large perturbation to the two largest entries and a relatively small perturbation to the other entries of the multinomial. 
\item {\bf Alternative Proportional to Null: } We perturb each entry of the null distribution by an amount proportional to $p_0(i)$, with a randomly chosen sign. 
\end{enumerate}

\begin{center}
\begin{figure}
\begin{tabular}{cc}
\includegraphics[scale=0.4]{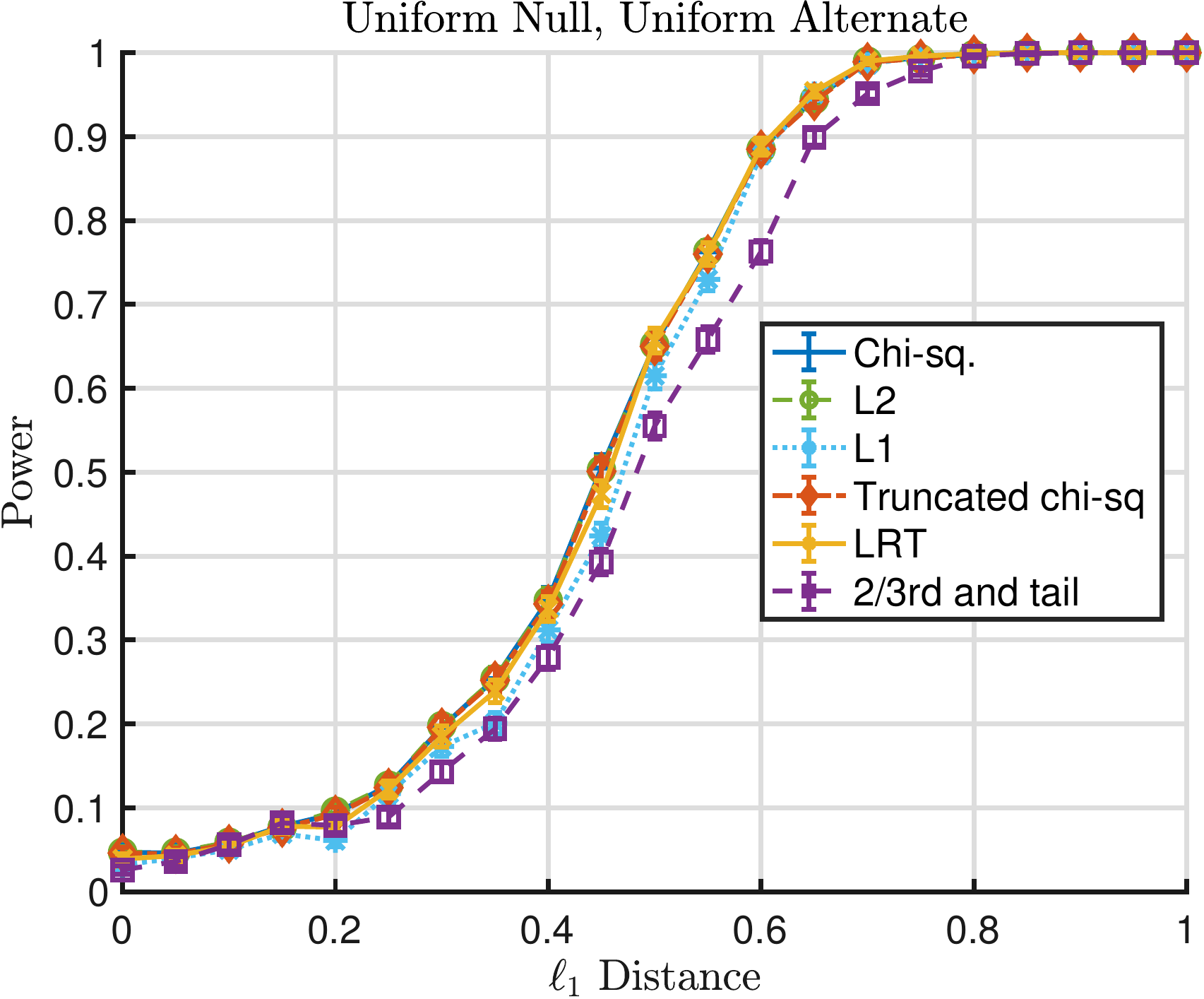} & ~~~~~\includegraphics[scale=0.4]{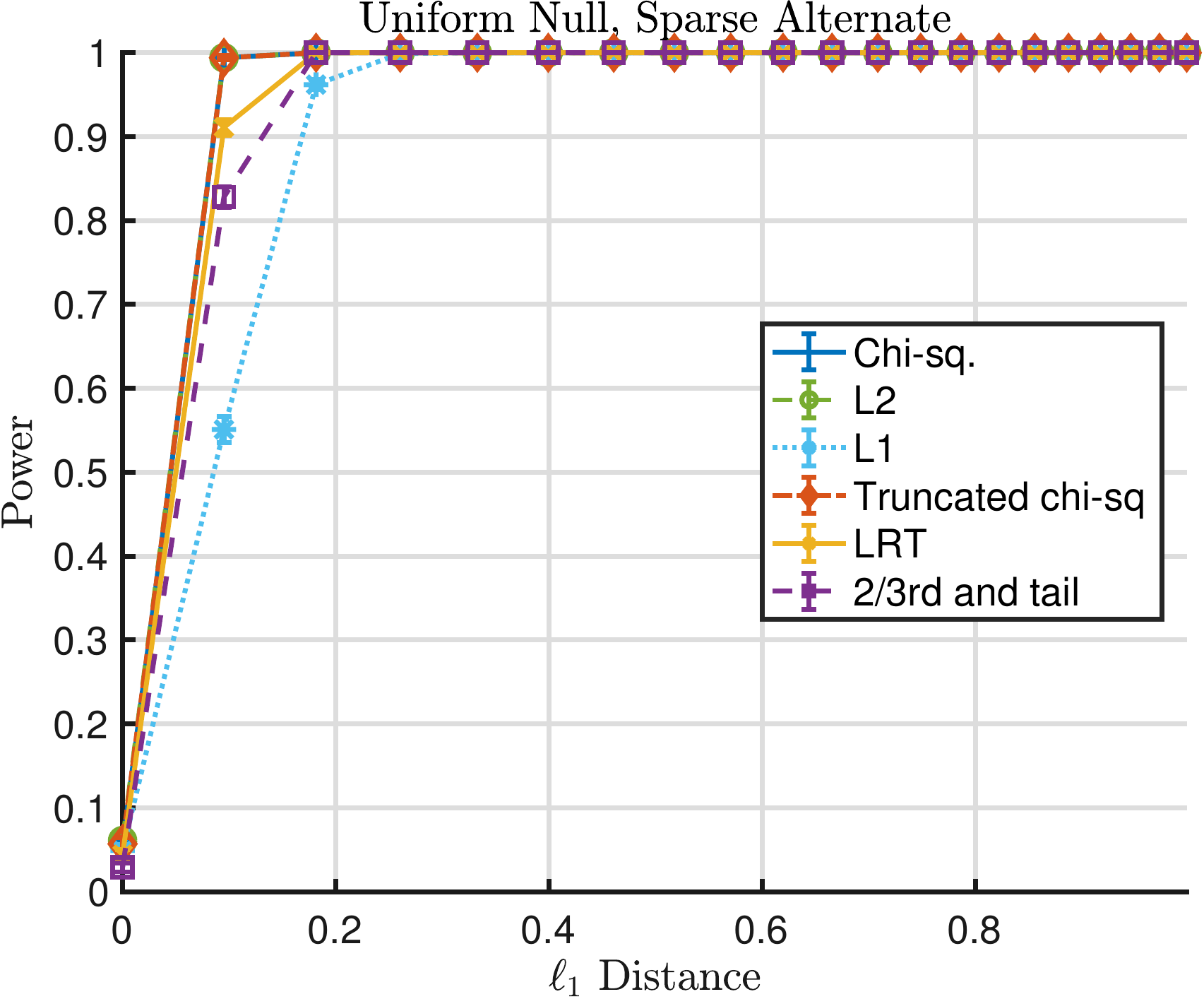}
\end{tabular}
\caption{A comparison between the truncated $\chi^2$ test, the 2/3rd + tail test \citep{valiant2017automatic}, 
the $\chi^2$-test, the likelihood ratio test, the $\ell_1$ test and the $\ell_2$ test. 
The null is chosen to be uniform, and the alternate is either a dense or sparse perturbation of the null. The power of the tests are plotted against the $\ell_1$ distance between the null and alternate. Each point in the graph is an average over 1000 trials. Despite the high-dimensionality (i.e. $n = 200, d = 2000$) the tests have high-power, and perform comparably.}
\label{fig:unifapp}
\end{figure}
\end{center}
We observe that the truncated $\chi^2$ test and the 2/3rd + tail test from \cite{valiant2017automatic} are remarkably robust. All tests are comparable when the null is uniform but the two-stage 2/3rd + tail test suffers a slight loss in power due to the Bonferroni correction. The distinctions between the classical tests and the recently proposed modified tests are clearer for the power law null.
In particular, from the simulation testing a power-law null against a sparse alternative it is clear that the $\chi^2$ and likelihood ratio test can have very poor power in the high-dimensional setting.
The $\ell_2$ test appears to have high-power against sparse alternatives but performs poorly against dense alternatives suggesting potential avenues for future investigation.

\begin{center}
\begin{figure}[h]
\begin{tabular}{cc}
\includegraphics[scale=0.4]{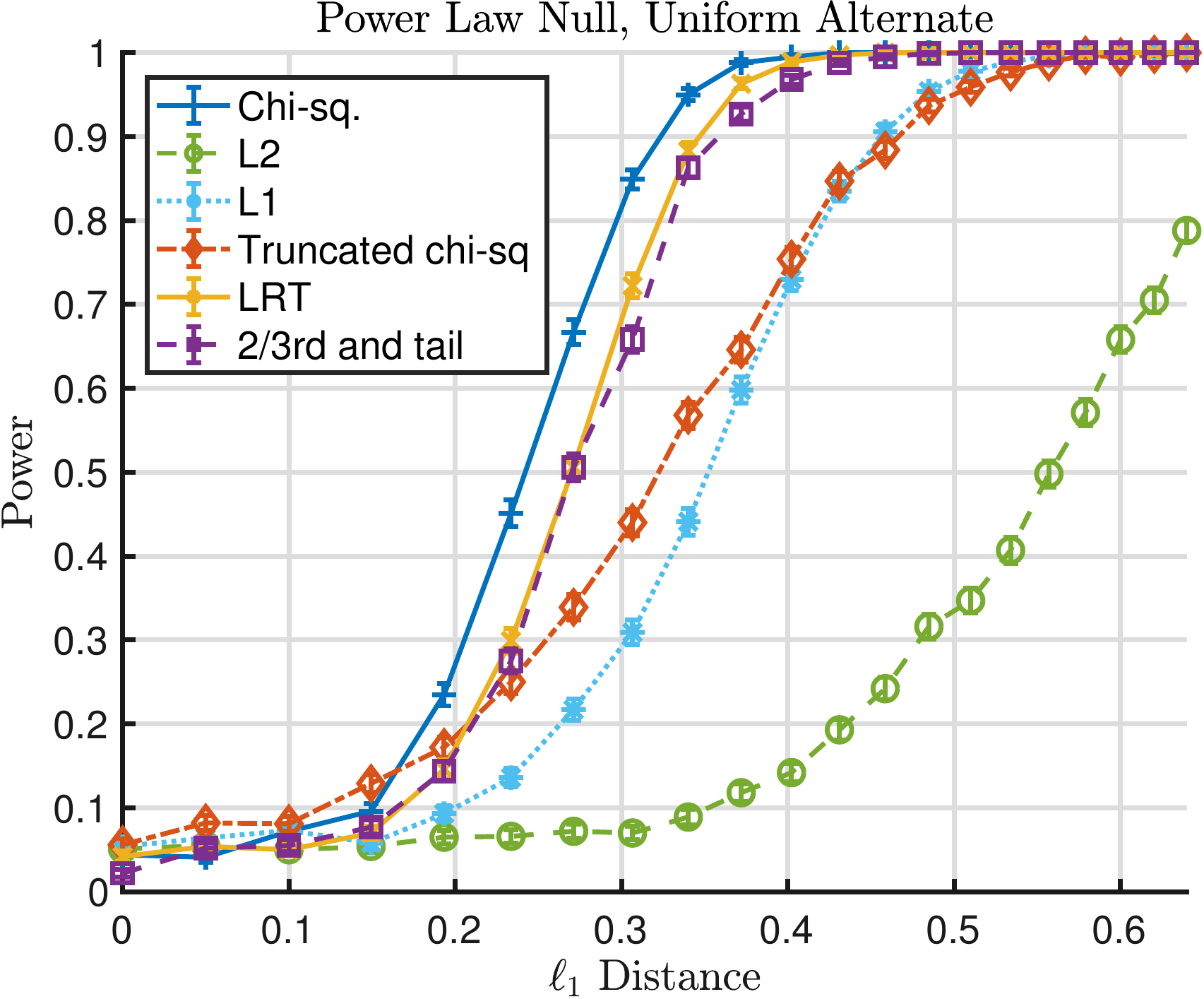} & ~~~~~\includegraphics[scale=0.4]{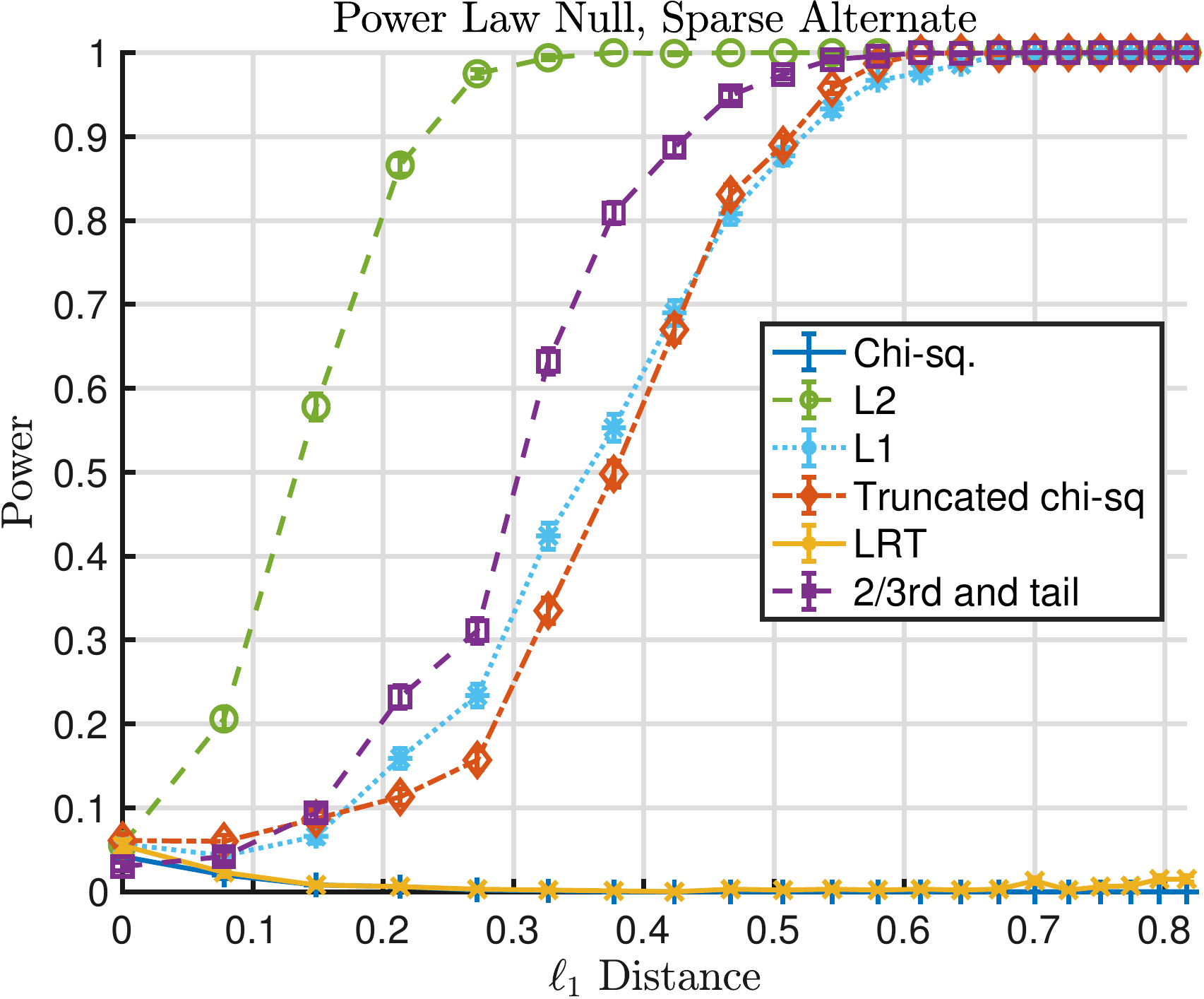} \\
\includegraphics[scale=0.4]{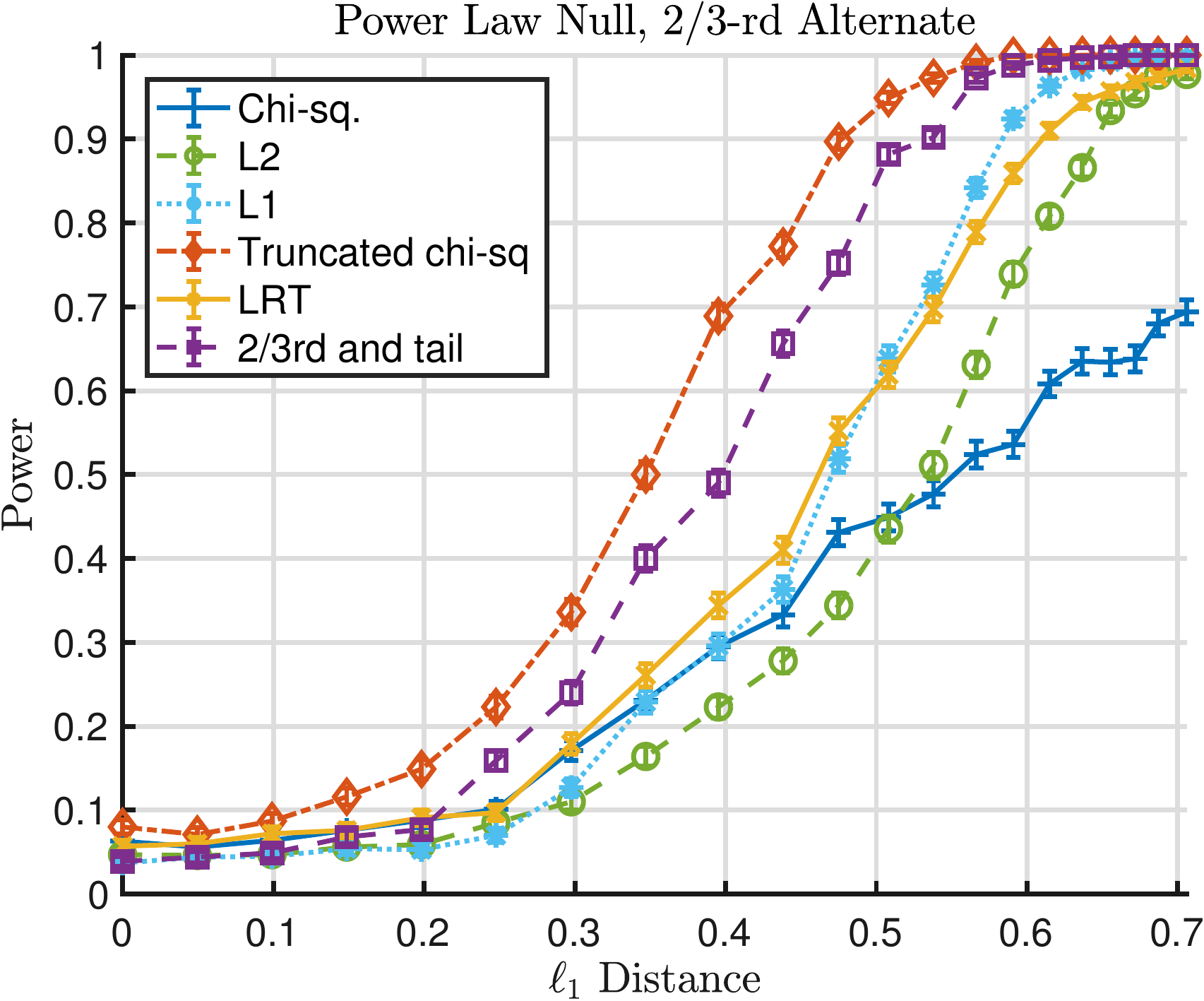} & ~~~~~\includegraphics[scale=0.4]{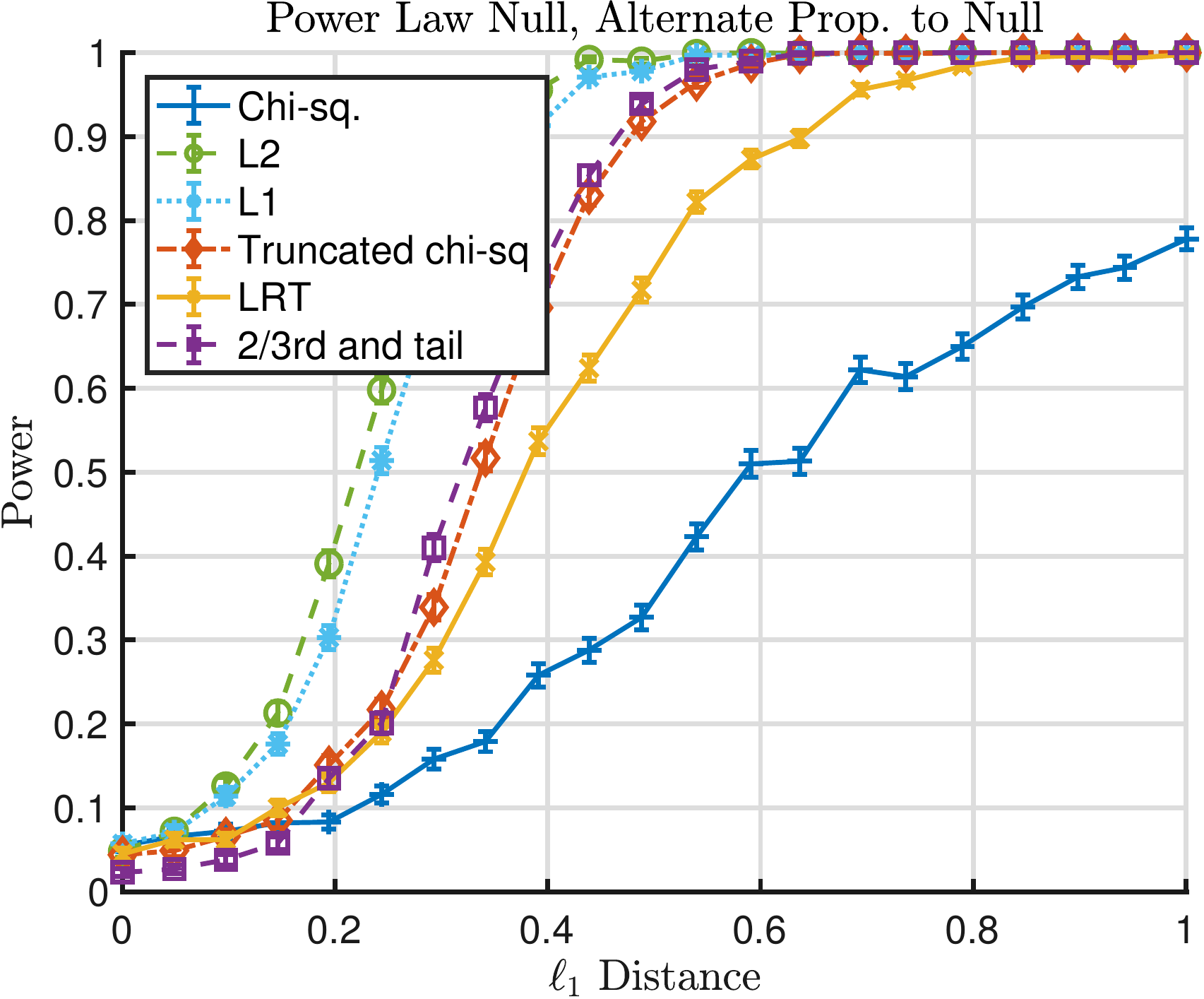} \\
\end{tabular}
\caption{A comparison between the truncated $\chi^2$ test, the 2/3rd + tail test \citep{valiant2017automatic}, 
the $\chi^2$-test, the likelihood ratio test, the $\ell_1$ test and the $\ell_2$ test. 
The null is chosen to be a power law with $p_0(i) \propto 1/i$.
We consider four possible alternates, the first uniformly perturbs the coordinates, the second
is a sparse perturbation only perturbing the first two coordinates, the third perturbs each co-ordinate proportional to $p_0(i)^{2/3}$ and the final setting perturbs each coordinate proportional to $p_0(i)$. The power of the tests are plotted against the $\ell_1$ distance between the null and alternate. Each point in the graph is an average over 1000 trials.}
\label{fig:powerapp}
\end{figure}
\end{center}

\subsection{Two-sample testing}

Finally, we turn our attention to the problem of two-sample testing
for high-dimensional multinomials.  We compare three different test
statistics, the two-sample $\chi^2$ statistic~\eqref{eqn:chisq_stat},
the $\ell_1$ and $\ell_2$ statistics:
\begin{align*}
T_{\ell_1} = \sum_{i=1}^d \left| \frac{X_i}{\ssone} - \frac{Y_i}{\sstwo} \right|,~~~~~~~~~
T_{\ell_2} = \sum_{i=1}^d \left(\frac{X_i}{\ssone} - \frac{Y_i}{\sstwo}\right)^2,
\end{align*}
and an oracle goodness-of-fit statistic. The oracle has access to the
distribution $P$ (and ignores the first sample) and tests
goodness-of-fit using the second sample. In our simulations, we use
the truncated $\chi^2$ test for goodness-of-fit.

\begin{center}
\begin{figure}[h]
\begin{tabular}{ccc}
\includegraphics[scale=0.4]{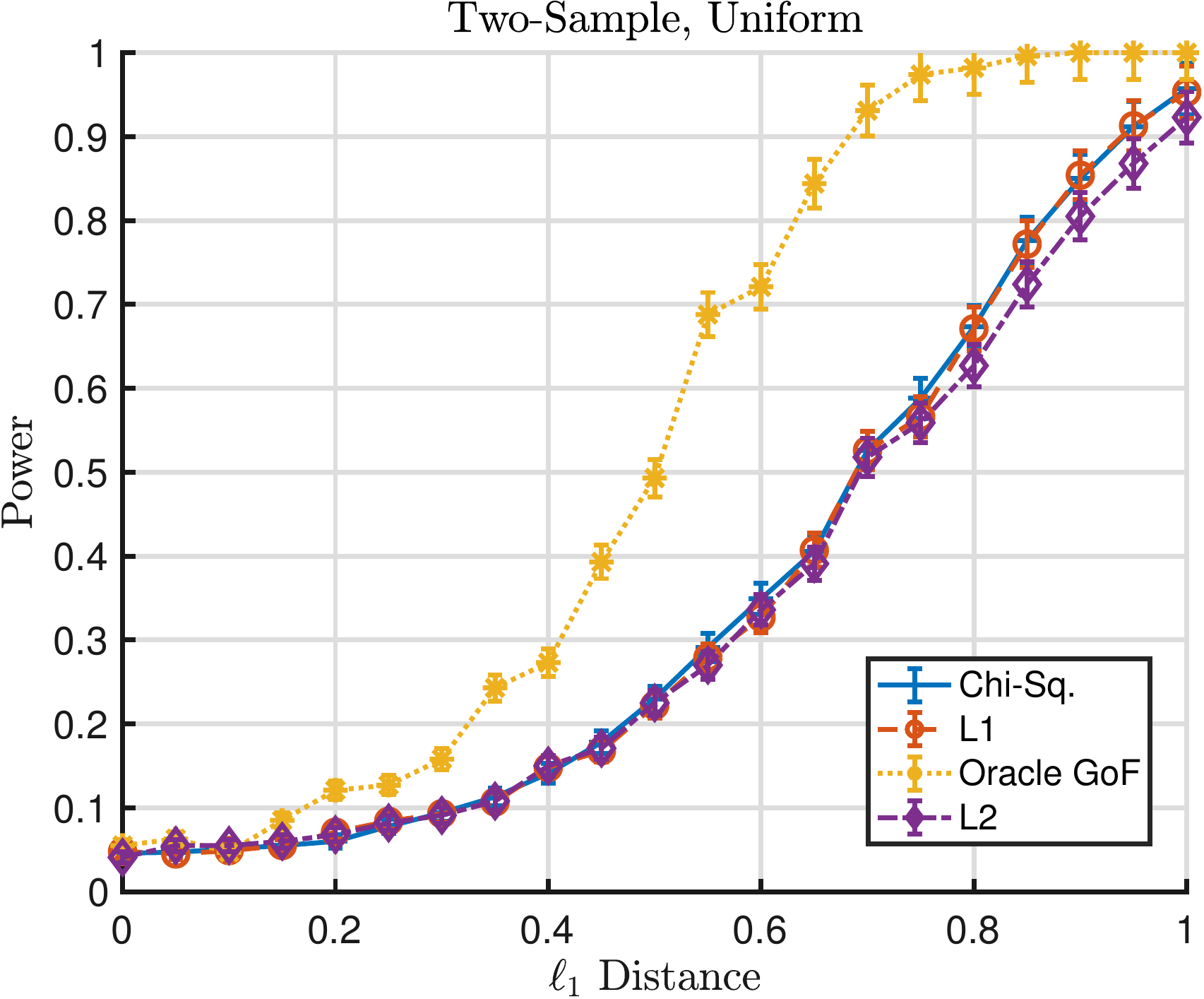} & ~~~~~\includegraphics[scale=0.4]{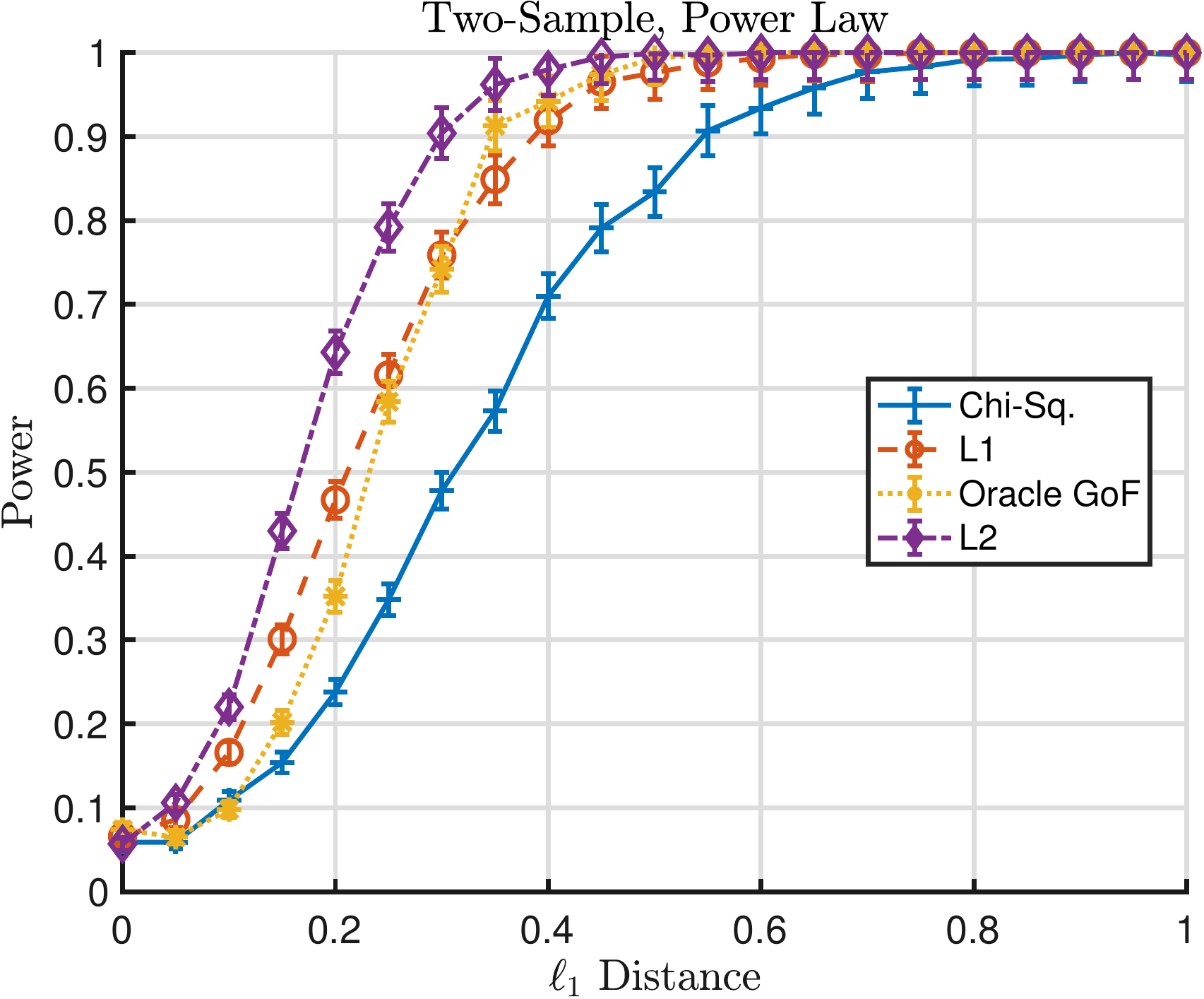} \\
\end{tabular}
\begin{center}
\begin{tabular}{c}
\includegraphics[scale=0.4]{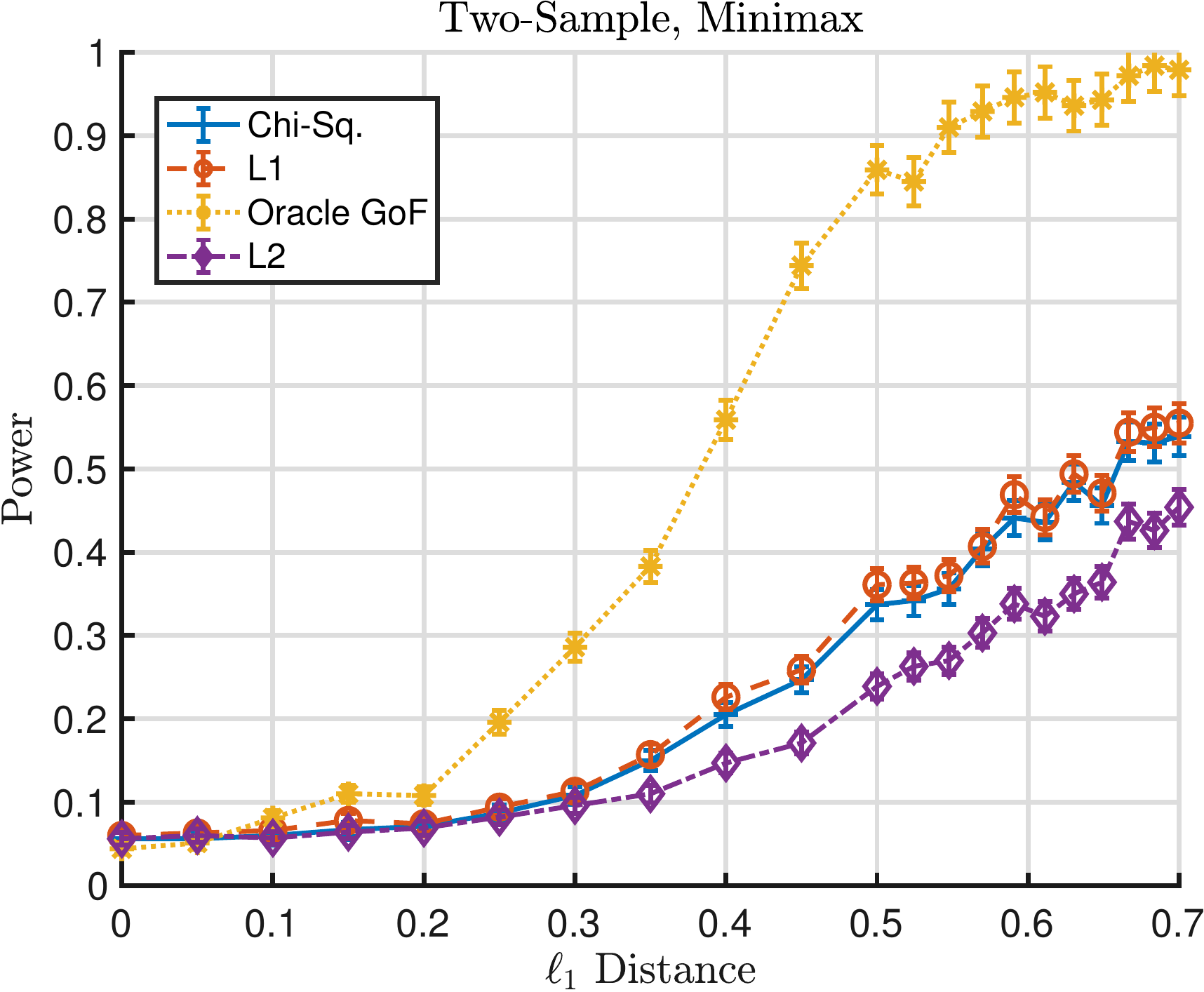} 
\end{tabular}
\end{center}
\caption{A comparison between the $\chi^2$ test, the $\ell_1$ test, the $\ell_2$ test, and 
the (oracle) goodness-of-fit test. In the three settings, the two distributions are chosen as described in the text (the distribution $P$ is chosen to be either uniform, power-law or minimax). 
The power of the tests are plotted against the $\ell_1$ distance between $P$ and $Q$. The sample sizes from $P$ and $Q$ are taken to be balanced and are each equal to $200$. Each point in the graph is an average over 1000 trials.}
\label{fig:balance}
\end{figure}
\end{center}

Motivated by our simulations in the previous section, we consider the following pairs of distributions
in the two-sample setup:
\begin{enumerate}
\item {\bf Uniform $P$, Dense Perturbation $Q$: } 
We take the distribution $P$ to be uniform and the distribution $Q$ to be the distribution where we perturb each entry of $P$ by a scaled Rademacher random variable.

\item {\bf Power-Law $P$, Sparse Perturbation $Q$: } Noting the difficulty faced by classical tests for goodness-of-fit testing of a power law versus a sparse perturbation (see Figure~\ref{fig:powerapp}) we consider a similar setup in the two-sample setting. We take each entry $p(i) \propto 1/i$ and take $q(i)$ to be the sparse perturbation described previously (where the two largest entries are perturbed by a relatively large magnitude).

\item {\bf Minimax $P, Q$: } This construction is inspired by the work of \cite{batu00}, and is used in their construction of a minimax lower bound for two-sample testing in the high-dimensional setting (i.e. when $d \gg \max\{\ssone, \sstwo\}$).

For a prescribed separation $\epsilon$, with probability $\ssone/2d$ we choose 
$p(i) = q(i) = 1/\ssone$ and with probability $1 - \ssone/2d$ we choose
$p(i) = 1/(2d)$ and $q(i) = 1/(2d) + \epsilon R_i/(2d)$, where the $R_i$ denote independent Rademacher random variables. Both distributions are then normalized.

Roughly, the two distributions contain a mixture of heavy elements of mass $1/\ssone$ and light elements of mass close to $1/d$. The two distributions have $\ell_1$ distance close to $\epsilon$ and the insight of \cite{batu00} is that in the two-sample 
setting is quite difficult to distinguish between variations in observed 
frequencies due to the perturbation of the entries and due to the random mixture of heavy and light entries.

\end{enumerate}
In each case, the cut-off for the tests is determined via the permutation method. 

\vspace{0.1cm}

{\noindent {\bf The balanced case: }} In this case, we set $\ssone = \sstwo = 200$ and $d = 400$. We observe in Figure~\ref{fig:balance} that there is a clear and significant loss in power relative to the goodness-of-fit oracle, indicating the increased complexity of two-sample testing in the high-dimensional setup. We note that, as is clear from the minimax separation rate we do not expect any loss in power in the low-dimensional setting. The loss in power is exacerbated when we consider the minimax two-sample pair $(P,Q)$ described above. We also note that the loss in power is negligible for the power law pair of distributions: due to the rapid decay of their entries these multinomials can be well estimated from a small number of samples.

\vspace{0.1cm}

\begin{center}
\begin{figure}[h]
\begin{tabular}{ccc}
\includegraphics[scale=0.4]{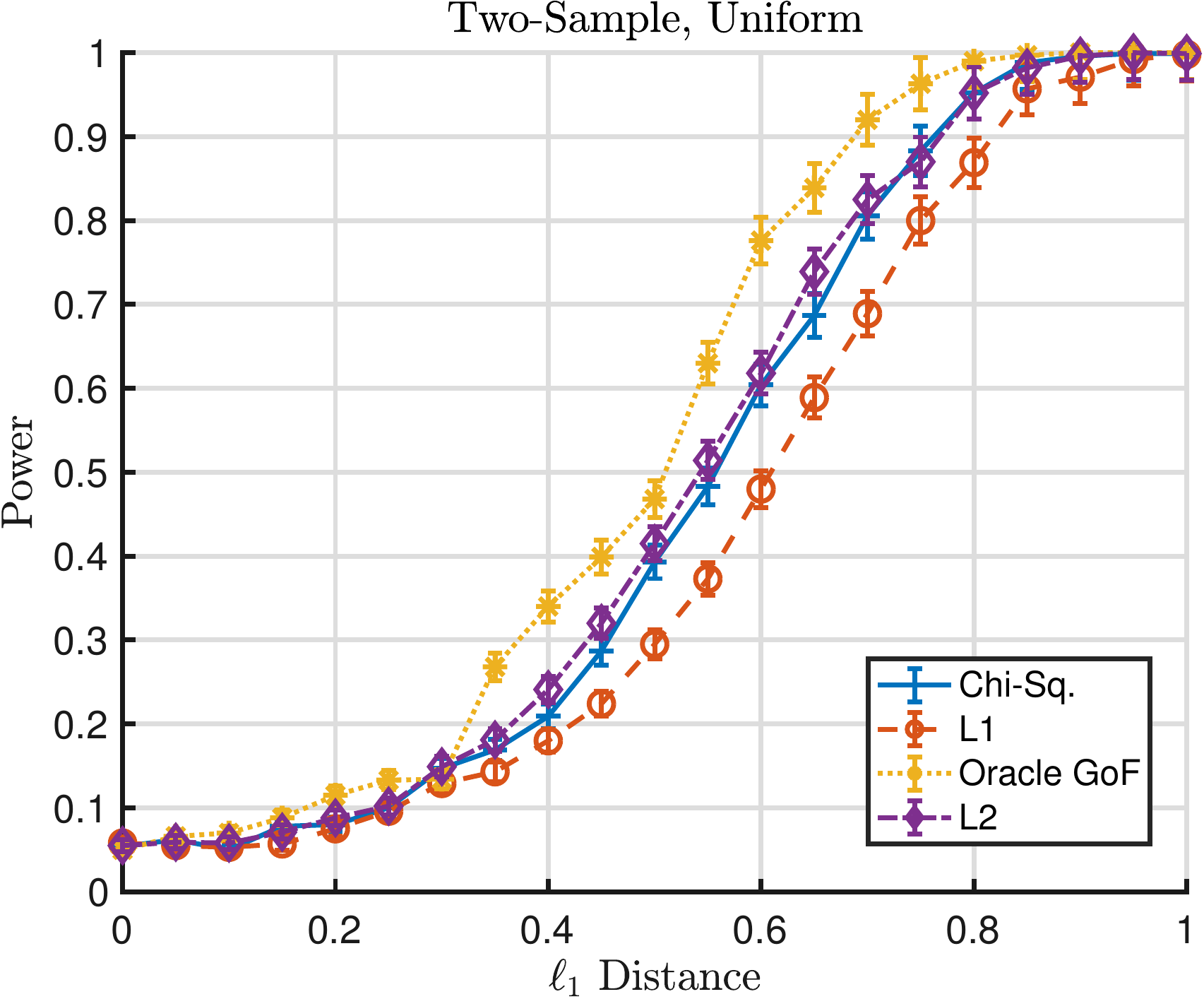} & ~~~~~\includegraphics[scale=0.4]{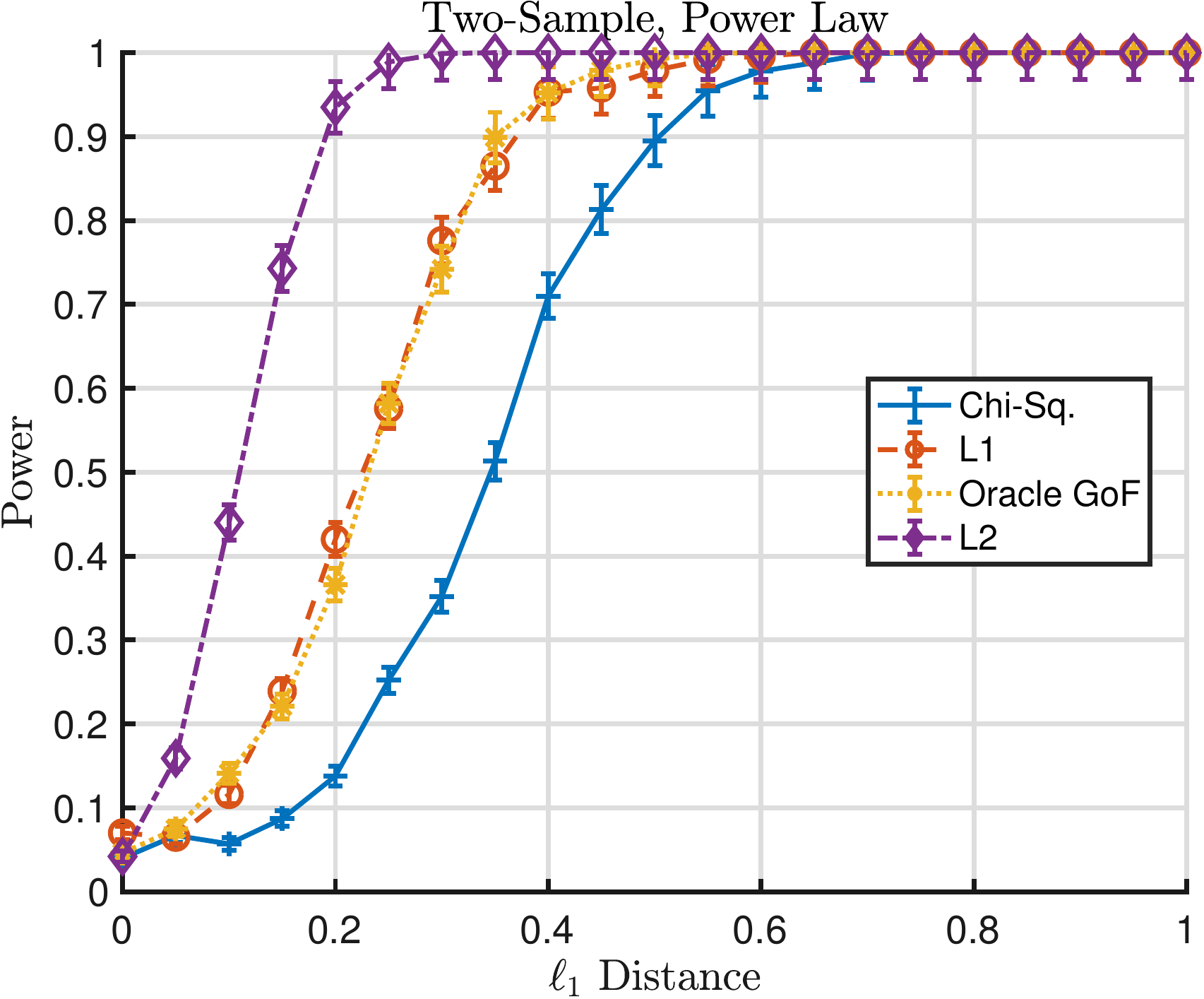} \\
\end{tabular}
\begin{center}
\begin{tabular}{c}
\includegraphics[scale=0.4]{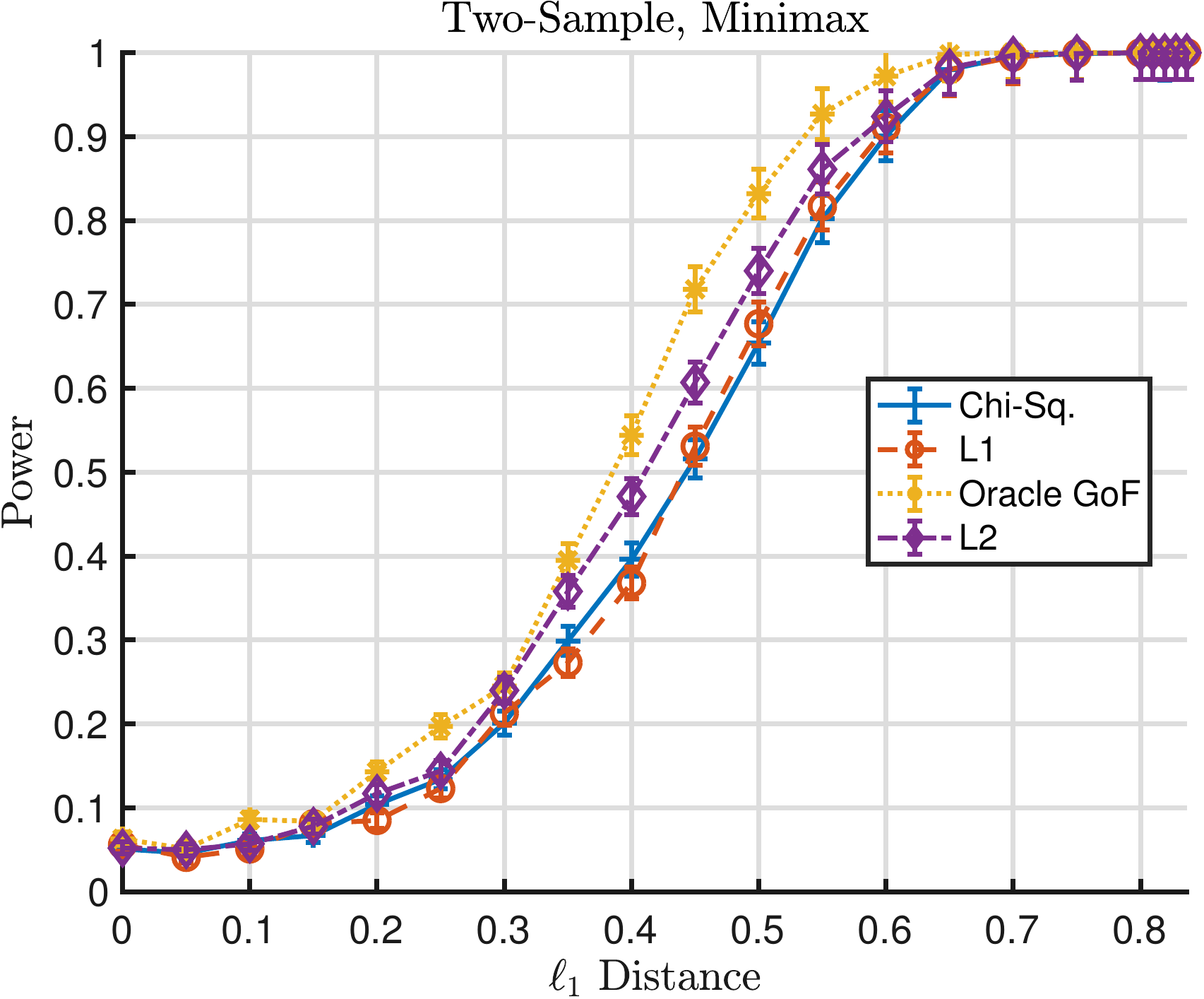} 
\end{tabular}
\end{center}
\caption{A comparison between the $\chi^2$ test, the $\ell_1$ test, the $\ell_2$ test, and 
the (oracle) goodness-of-fit test. In the three settings, the two distributions are chosen as described in the text (the distribution $P$ is chosen to be either uniform, power-law or minimax). 
The power of the tests are plotted against the $\ell_1$ distance between $P$ and $Q$. The sample sizes from $P$ and $Q$ are taken to be imbalanced, i.e. we take $\ssone = 2000$ and $\sstwo = 200$. Each point in the graph is an average over 1000 trials.}
\label{fig:imbalance}
\end{figure}
\end{center}

{\noindent {\bf The imbalanced case: }} In this case, we set $\ssone = 2000, \sstwo = 200$ and $d = 400.$ The $\chi^2$ statistic for two-sample testing in the imbalanced case, is given by: 
\begin{align*}
T = \sum_{i=1}^d \frac{(\sstwo X_i - \ssone Y_i)^2 - \sstwo^2 X_i - \ssone^2 Y_i}{X_i + Y_i}
\end{align*}
where we follow \cite{bhattacharya2015testing} and use a slightly modified centering of the usual $\chi^2$ statistic. Since the $\chi^2$-statistic is minimax optimal in the balanced case, one might conjecture that this continues to be the case in the imbalanced case. However, as our simulations suggest this is not the case. Somewhat surprisingly, the performance of the $\chi^2$ statistic can degrade when one of the sample sizes is increased (see Figure~\ref{fig:imbalance}). The $\ell_1$ statistic on the other hand appears to be perform as expected, i.e. its power is close to that of the oracle goodness-of-fit test in the case when one of the sample sizes is very large, and we believe that this statistic warrants further study. 
%In \cite{bhattacharya2015testing} a combination of the $\chi^2$ and $\ell_1$ test are used to achieve near-minimax rates.  

\section{Discussion}
\label{sec:discussion}

Despite the fact that discrete data analysis is an old subject,
it is still a vibrant area of research
and there is still much that we don't know.
Steve Fienberg showed prescience in drawing attention to
one of the thorniest issues: high-dimensional multinomials.

Much of the statistical literature has dealt with the high-dimensional case by
imposing assumptions on the distribution so that
simple limiting distributions can be obtained.
Doing so gives up the most appealing property of multinomial inference:
it is completely distribution-free.
As we have seen in this paper, recent work by a variety of communities has
developed new and rather surprising theoretical results.
What is often missing in the recent literature is the appreciation that statisticians
want tests with precise control of the type I error rate.
As a result, there remain gaps between theory and practice.

We have focused on goodness of fit and two sample problems.
There is a rich literature on other problems
such as independence testing and testing shape constraints
\citep{diakonikolas2016new,acharya15}.
As we discussed earlier,
\cite{balakrishnan2017hypothesis}
showed that these new results for high-dimensional discrete data
have implications for continuous data.
There is much more to say about this and
this is a direction that we are actively pursuing.
Finally, we have restricted attention to hypothesis testing.
In future work, we will report results
on high dimensional inference using
confidence sets and point estimation.
%\section{to be added}
%
%Cressie and Read, Holst (1972) already suggest unnatural to keep $d$
%fixed. 
%
%Hoeffding (1965) points out that LRT may no longer dominate
%$\chi^2$. Morris gave CLTs.
%%\usepackage{natbib}
%
%Local power analysis, uniformly optimal tests do not exist (Ivchenko
%and Medvedev 1978). What statistic to use if $d$ grows and
%non-uniform?
%
%Minimax!
%
%
%
%Connections to work of Candes, El Karoui, Donoho -- asymptotic predictions in high-dimensions?

\section{Acknowledgements}
This research was supported by NSF grant DMS1713003. All of our simulation results and figures can be reproduced using code made available at (\url{http://www.stat.cmu.edu/~siva/multinomial}).

\bibliographystyle{siva}
\bibliography{paper}

\section*{Appendix}

Here we describe the local minimax results for goodness of fit testing
more precisely.
Without loss of generality 
we assume that the entries of the null multinomial $p_0$ 
are sorted so that $p_0(1) \geq p_0(2) \geq \ldots 
\geq p_0(d)$. For any $0 \leq \sigma \leq 1$ we denote $\sigma$-tail of the
multinomial by:
\begin{align}
\label{eqn:taildef}
\epstail{\sigma} = \left\{i: \sum_{j = i}^d p_0(j) \leq \sigma \right\}.
\end{align}
The $\sigma$-bulk is defined to be
\begin{align}
\label{eqn:bulkdef}
\bulk{\sigma} = \{i>1:\ i\notin \epstail{\sigma}\}.
\end{align}
Note that $i=1$ is excluded from the $\sigma$-bulk.
The minimax rate depends on the functional:
\begin{align}
\label{eqn:vfunc}
\truncnorm{\sigma}{p_0} = 
\left( \sum_{i \in \bulk{\sigma}} p_0(i)^{2/3} \right)^{3/2}.
\end{align}
Define, $\ell_n$ and $u_n$ to be the solutions to the equations:
\begin{align}
\label{eqn:multrad}
\ell_n(p_0) = \max\left\{ \frac{1}{n}, \sqrt{\frac{ \truncnorm{\ell_n(p_0)}{p_0}}{n}} \right\},\ \ \ 
u_n(p_0) = \max\left\{\frac{1}{n}, \sqrt{\frac{ \truncnorm{u_n(p_0)/16}{p_0}}{n}}\right\}.
\end{align}
With these definitions in place, we are now ready to state the result of
\cite{valiant2017automatic}.
We use $c_1,c_2,C_1,C_2 > 0$ to denote positive universal constants.

\begin{theorem}[\cite{valiant2017automatic}]
\label{thm:val-val}
The local critical radius $\epsilon_n(p_0,{\cal M})$ 
for multinomial testing is upper and lower bounded as:
\begin{align}
\label{eqn:minimaxcritrad}
c_1 \ell_n(p_0) \leq \epsilon_n(p_0,{\cal M}) \leq C_1 u_n(p_0).
\end{align}
Furthermore, the global critical radius $\epsilon_n({\cal M})$ is bounded as:
\begin{align*}
\frac{c_2 d^{1/4}}{\sqrt{n}} \leq \epsilon_n({\cal M}) \leq  \frac{C_2 d^{1/4}}{\sqrt{n}}.
\end{align*}
\end{theorem}

\end{document}